\documentclass[journal]{IEEEtran}
\usepackage[linesnumbered,ruled]{algorithm2e}
\usepackage{graphicx,amssymb,mathrsfs,amsmath,array,color,amsthm}
\usepackage{subfigure}
\usepackage{multirow}
\usepackage{booktabs}
\usepackage{cite}
\usepackage[colorlinks,urlcolor=blue,driverfallback=dvipdfm]{hyperref}

\newcommand{\norm}[1]{\lVert#1\rVert}
\DeclareMathOperator{\F}{F}

\begin{document}
\title{More Diverse Means Better: Multimodal Deep Learning Meets Remote Sensing Imagery Classification}

\author{Danfeng Hong,~\IEEEmembership{Member,~IEEE,}
        Lianru Gao,~\IEEEmembership{Senior Member,~IEEE,}
        Naoto Yokoya,~\IEEEmembership{Member,~IEEE,}
        Jing Yao,
        Jocelyn Chanussot,~\IEEEmembership{Fellow,~IEEE,}
        Qian Du,~\IEEEmembership{Fellow,~IEEE,}
        and~Bing Zhang,~\IEEEmembership{Fellow,~IEEE}

\thanks{This work was supported by the National Natural Science Foundation of China under Grant 41722108 and 91638201 and the Japan Society for the Promotion of Science (JSPS) under Grant KAKENHI 18K18067, and with the support of the AXA Research Fund. (\emph{Corresponding author: Lianru Gao}).}
\thanks{D. Hong is with the Univ. Grenoble Alpes, CNRS, Grenoble INP, GIPSA-lab, 38000 Grenoble, France. (e-mail: hongdanfeng1989@gmail.com)}
\thanks{L. Gao is with the Key Laboratory of Digital Earth Science, Aerospace Information Research Institute, Chinese Academy of Sciences, 100094 Beijing, China. (e-mail: gaolr@aircas.ac.cn)}
\thanks{N. Yokoya is with Graduate School of Frontier Sciences, the University of Tokyo, 277-8561 Chiba, Japan, and also with the Geoinformatics Unit, RIKEN Center for Advanced Intelligence Project (AIP), RIKEN, 103-0027 Tokyo, Japan. (e-mail: naoto.yokoya@riken.jp)}
\thanks{J. Yao is with the School of Mathematics and Statistics, Xi’an Jiaotong University, 710049 Xi’an, China. (e-mail: jasonyao@stu.xjtu.edu.cn)}
\thanks{J. Chanussot is with the Univ. Grenoble Alpes, INRIA, CNRS, Grenoble INP, LJK, 38000 Grenoble, France, and also with the Aerospace Information Research Institute, Chinese Academy of Sciences, 100094 Beijing, China. (e-mail: jocelyn@hi.is)}
\thanks{Q. Du is with the Department of Electrical and Computer Engineering, Mississippi State University, Starkville, 39762 MS, USA. (e-mail: du@ece.msstate.edu)}
\thanks{B. Zhang is with the Key Laboratory of Digital Earth Science, Aerospace Information Research Institute, Chinese Academy of Sciences, 100094 Beijing, China, and also with the College of Resources and Environment, University of Chinese Academy of Sciences, 100049 Beijing, China. (e-mail: zb@radi.ac.cn)}
}

\markboth{IEEE Transactions on Geoscience and Remote Sensing,~Vol.~XX, No.~XX, ~XXXX,~2020}
{Shell \MakeLowercase{\textit{et al.}}: More Diverse Means Better: Multimodal Deep Learning Meets Remote Sensing Imagery Classification}

\maketitle
\begin{abstract}
\textcolor{blue}{This is the pre-acceptance version, to read the final version please go to IEEE Transactions on Geoscience and Remote Sensing on IEEE Xplore.} Classification and identification of the materials lying over or beneath the Earth's surface have long been a fundamental but challenging research topic in geoscience and remote sensing (RS) and have garnered a growing concern owing to the recent advancements of deep learning techniques. Although deep networks have been successfully applied in single-modality-dominated classification tasks, yet their performance inevitably meets the bottleneck in complex scenes that need to be finely classified, due to the limitation of information diversity. In this work, we provide a baseline solution to the aforementioned difficulty by developing a general multimodal deep learning (MDL) framework. In particular, we also investigate a special case of multi-modality learning (MML) -- cross-modality learning (CML) that exists widely in RS image classification applications. By focusing on ``what'', ``where'', and ``how'' to fuse, we show different fusion strategies as well as how to train deep networks and build the network architecture. Specifically, five fusion architectures are introduced and developed, further being unified in our MDL framework. More significantly, our framework is not only limited to pixel-wise classification tasks but also applicable to spatial information modeling with convolutional neural networks (CNNs). To validate the effectiveness and superiority of the MDL framework, extensive experiments related to the settings of MML and CML are conducted on two different multimodal RS datasets. Furthermore, the codes and datasets will be available at: \url{https://github.com/danfenghong/IEEE_TGRS_MDL-RS}, contributing to the RS community.
\end{abstract}
\graphicspath{{figures/}}


\section{Introduction}
\IEEEPARstart{B}{eyond} any doubt, remotely sensed image classification or mapping \cite{chanussot2006classification,hong2017learning,fauvel2012advances,cao2020an,li2019deep,hong2019learning,cao2020hyperspectral,rasti2020feature}, i.e., land use and land cover (LULC), plays an increasingly significant role in earth observation (EO) missions, as many high-level applications, to a great extent, depend on classification products, such as urban development and planning, forest monitoring, soil composition analysis, disaster response and management, to name a few.

\begin{figure}[!t]
	  \centering
		\subfigure{
			\includegraphics[width=0.48\textwidth]{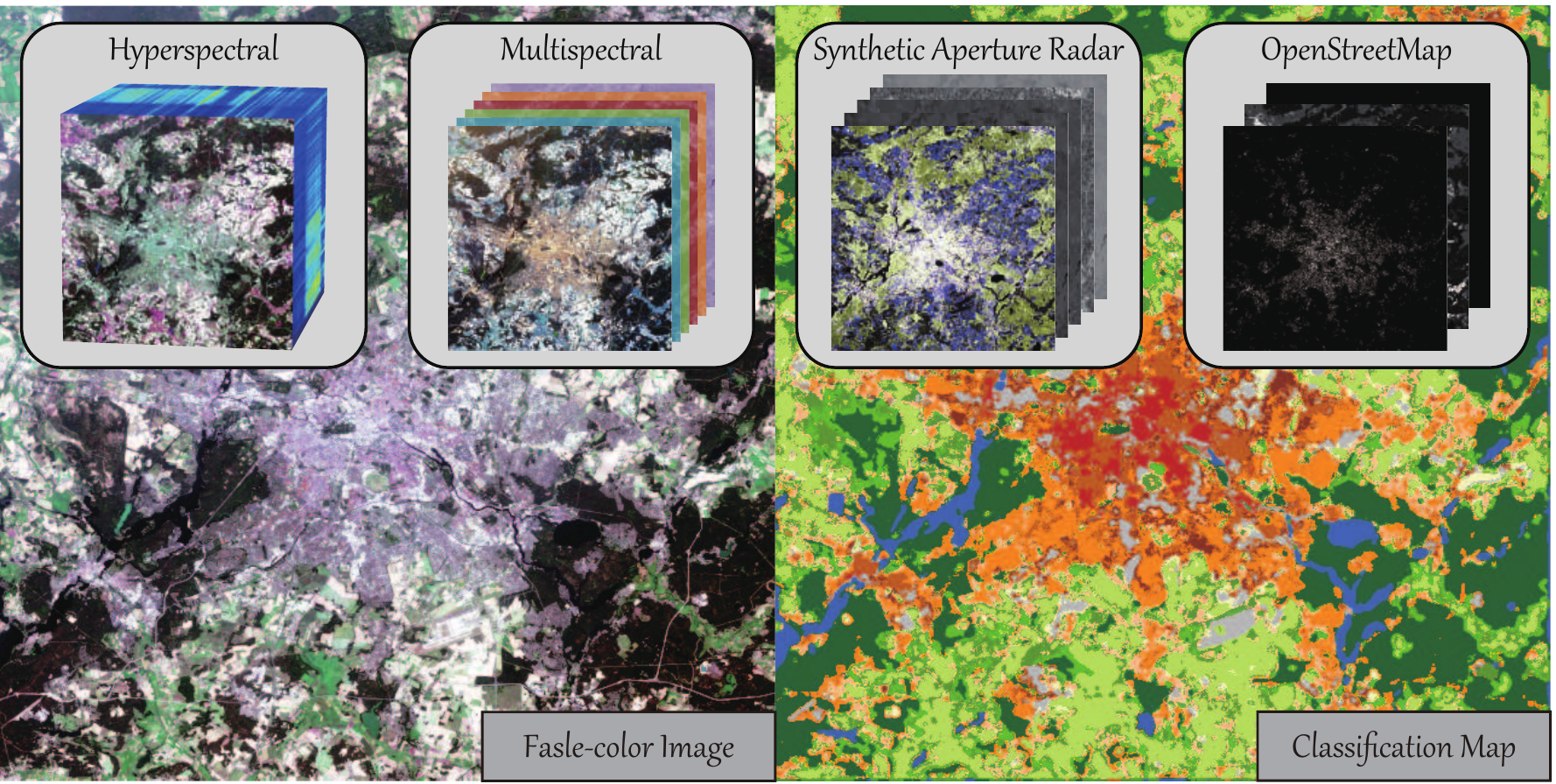}
		}
        \caption{A multimodal example (\textit{Berlin}) for RS imagery classification, where four data sources, i.e., HS, MS, SAR, and OSM, are available in a same scene.}
\label{fig:Example_MM}
\end{figure}

\begin{figure*}[!t]
	  \centering
		\subfigure[Training flow for MML and CML]{
			\includegraphics[width=0.315\textwidth]{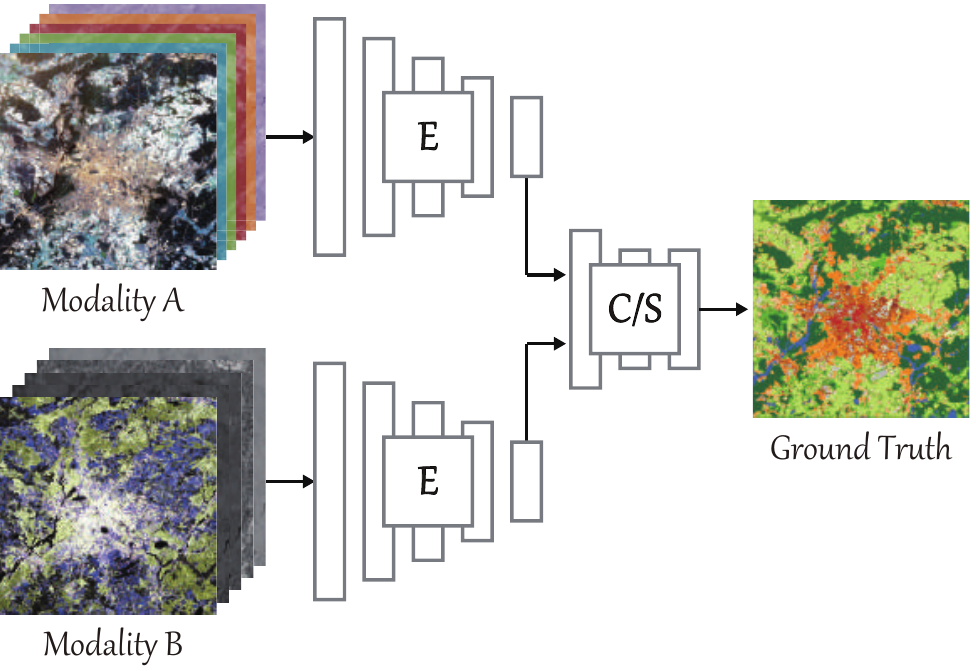}
            \label{fig:MML_TR}
		}
		\subfigure[Testing flow for MML]{
			\includegraphics[width=0.315\textwidth]{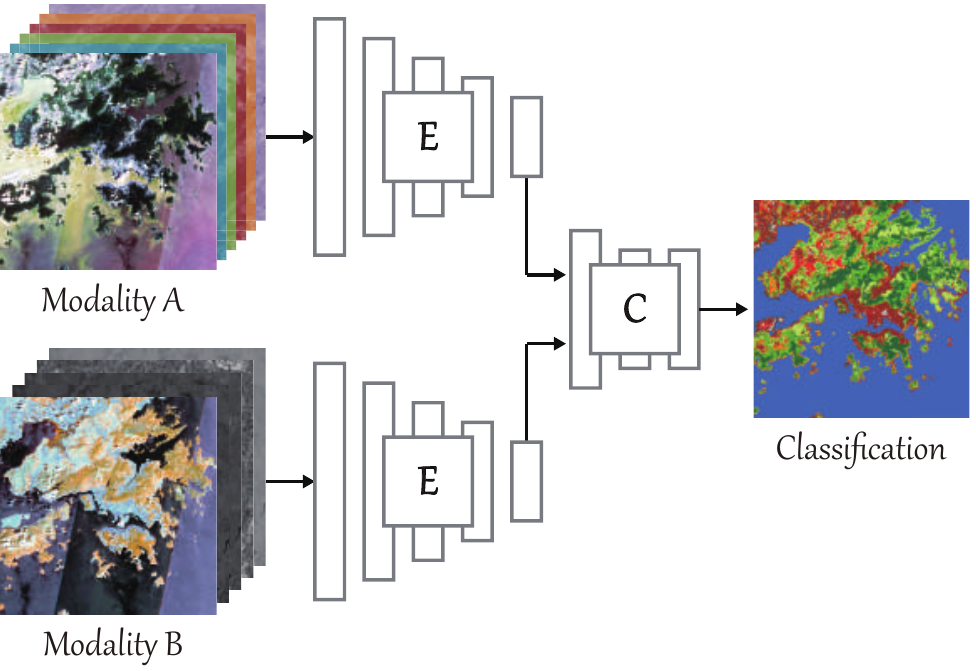}
            \label{fig:MML_TE}
		}
	    \subfigure[Testing flow for CML]{
			\includegraphics[width=0.315\textwidth]{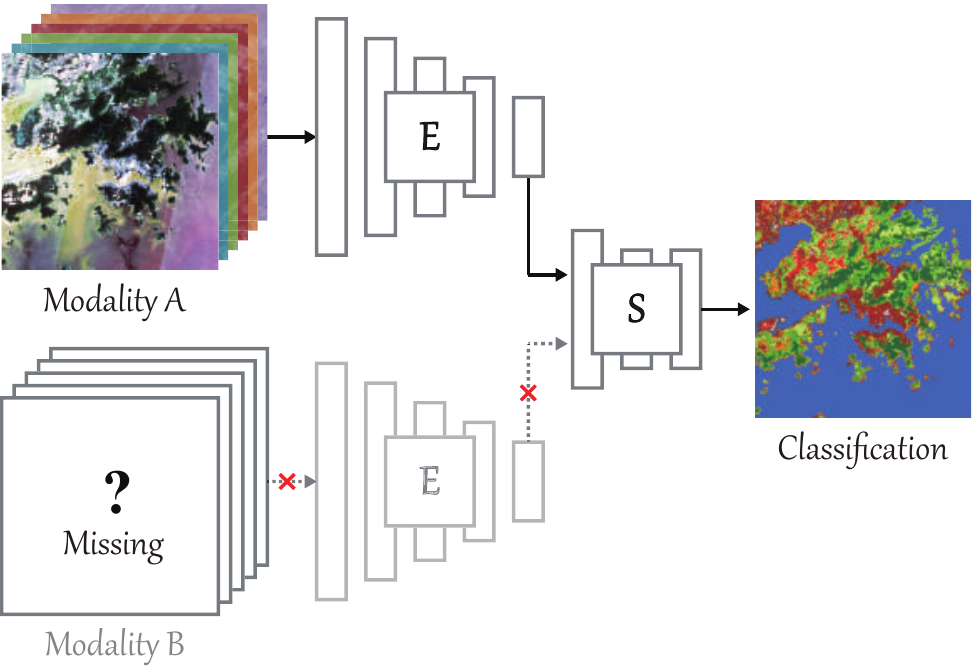}
            \label{fig:CML_TE}
		}
         \caption{An illustration to clarify the similarities and differences between MML and CML from training and testing perspectives. Note that MML and CML aim to learn the model over multiple modalities in the training phase, e.g., feature extractor (E) and feature fusion (C: concatenated and S: shared). The main difference lies in that one modality is absent for CML during inference time. In other words, certain modality is not involved in the prediction phase.}
\label{fig:MML_CML}
\end{figure*}

Over the past decades, enormous effects have been made to extract discriminative features and design efficient classifiers for remote sensing (RS) data classification. However, most of these classification techniques, either unsupervised or supervised, are merely designed and applied for single modalities, e.g., hyperspectral (HS) \cite{yao2019nonconvex}, multispectral (MS) \cite{wu2019fourier}, light detection and ranging (LiDAR) \cite{huang2019multi}, synthetic aperture radar (SAR) \cite{kang2020learning}, OpenStreetMap (OSM), etc. The ability in identifying materials on the surface of the Earth, therefore, remains limited, due to the lack of rich and diverse information, particularly in challenging scenes where certain categories are similar and cannot be accurately classified by only single modalities. For instance, in urban planning, the structure types of surface materials are hardly identified using only one modality information (e.g., spectral data) \cite{hong2018augmented}. There are no big differences in spectral profiles between the ``grass'' on the ground and the ``grass'' on the roof, but they can be well separated by means of height information obtained from lidar or SAR data \cite{heiden2012urban}. In object detection and localization (e.g., cars), the HS data is characterized by more discriminative spectral properties, while the RGB or MS products are capable of providing richer and finer spatial information \cite{mayer2003object}. This is also a typical win-win case. Moreover, it is well known that optical RS images suffer from the effects of cloud coverage in image acquisition, leading to partial information missing. SAR can be seen as an auxiliary data source to address the issue effectively, due to its different imaging mechanism and sensors that are able to penetrate the cloud \cite{huang2015cloud}.

Different imaging technologies in RS are capable of capturing a variety of properties from the Earth's surface, such as spectral radiance and reflectance, height information, texture structure, and spatial characteristics. The joint exploitation of multiple modalities enables us to characterize the scene at a more detailed and precise level unachievable by using single modality data \cite{gomez2015multimodal}. In addition, a large amount of multimodal earth observation data, such SAR, MS, HS, and digital surface model (DSM), become openly available from currently operational spaceborne radar (e.g., Sentinel-1), optical broadband (e.g., Sentinel-2, Landsat-8), and imaging spectroscopy (e.g., Hyperion, DESIS, Gaofen-5) missions as well as various airborne sensors (e.g., HyMap, HySpex) or laser scanning \cite{wehr1999airborne}. Fig. \ref{fig:Example_MM} shows a classification example with multimodal RS data. This further motivates us to investigate and design advanced multimodal data analysis (MDA) techniques. Despite many conventional MDA-related approaches proposed and used by attempts to enhance the classification results of RS data sources, yet the relatively poor capability of these models in data representation limit the performance gain \cite{liao2014generalized,ghamisi2015land,yokoya2018open,gao2020spectral}. Inspired by the recent success of deep learning (DL), some preliminary studies \cite{chen2017deep,xu2017multisource,benedetti2018m,hang2019cascaded} have addressed this issue with the multimodal input. Their outcomes, to some extent, have shown a great potential in RS imagery classification tasks.

Nevertheless, there still lacks a unified MDA-targeted DL architecture that is able to clarify three open questions, that is, ``what to fuse'', ``how to fuse'', and ``where to fuse''. To this end, we propose a general multimodal DL framework for the RS imagery classification. The proposed model aims to provide an inclusive baseline network to break the bottleneck of classification performance under the conditions of using single modalities, where several fusion modules can be well embedded. Furthermore, extensive experiments conducted on three different multi-modal datasets freely available demonstrate the MDL-RS's superiority in terms of either the common multi-modality learning (MML) or the special cross-modality learning (CML) issue (see Fig. \ref{fig:MML_CML}) using the fully connected (FC) networks (FC-Nets for short) and convolutional neural networks (CNNs). The main contributions in this paper can be highlighted as follows.

\begin{itemize}
\item We propose a unified multimodal DL framework with a focus on the RS image classification, MDL-RS for short, which assembles pixel-level labeling guided by an FC design and spatial-spectral joint classification with CNNs-dominated architecture.
\item The proposed MDL-RS is not only applicable to the case of MML, but also able to be generalized to CML's with more effective and compact modality blending.
\item Five plug-and-play fusion modules are investigated and devised in the MDL-RS networks. They are \textit{early fusion}, \textit{middle fusion}, \textit{late fusion}, \textit{encoder-decoder (En-De) fusion}, and \textit{cross fusion}, where the first four approaches are the well-known fusion strategy yet lack of being generalized in a unified framework, and the last one is a newly-proposed contribution that can transfer the information across modalities more effectively.
\end{itemize}


\begin{figure*}[!t]
	  \centering
		\subfigure{
			\includegraphics[width=1\textwidth]{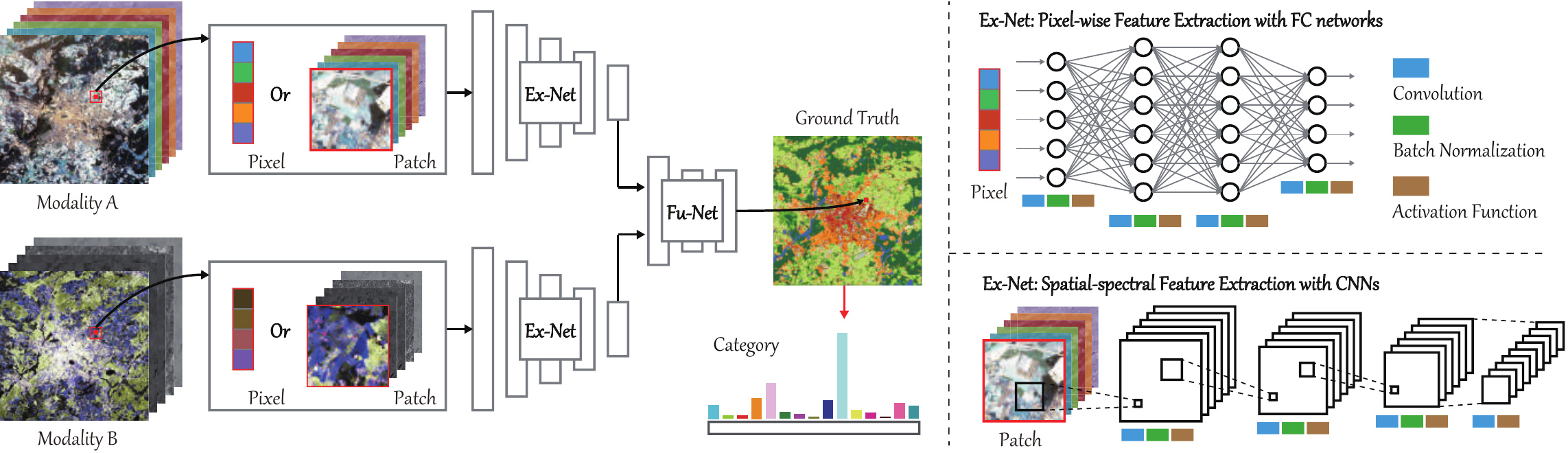}
		}
        \caption{An illustration overview of our proposed MDL-RS framework for the RS imagery classification with two subnetworks: \textit{Extraction Network (Ex-Net)} and \textit{Fusion Network (Fu-Net)}. This \textit{Ex-Net} in MDL-RS consists of two different feature extractors: pixel-wise FC-Nets and spatial-spectral CNNs.}
\label{fig:workflow_MDL}
\end{figure*}

\section{Related Work}
Fig. \ref{fig:MML_CML} briefly illustrates the MML and CML for training and testing. Accordingly, we will highlight some significant works related to the two topics in the following.

\subsection{Shallow Models for MML}
Many classic shallow models related to MML, i.e., morphological operators and subspace learning, have been successfully employed for feature extraction and classification of multimodal RS observations. For example, Liao \textit{et al.} \cite{liao2014generalized} proposed to fuse the morphological profiles (MPs) of HS and LiDAR data on manifolds by means of graph-based subspace learning. Similarly, Ref. \cite{ghamisi2015land} extracted the attribute profiles (APs) instead of MPs used in \cite{liao2014generalized} for land-cover classification. In \cite{rasti2017hyperspectral}, extinction profiles (EPs) combined with total variation component analysis are used for the fusion of HS and LiDAR. The fusion work is improved in \cite{rasti2017fusion} by jointly using sparse and low-rank subspace modeling. Yokoya \textit{et al.} \cite{yokoya2018open} simply stacked multiple features obtained from MS and OpenStreetMap data before feeding into the classifier for local climate zones (LCZ) classification. Supported by topological theory, Hu \textit{et al.} \cite{hu2019mima} developed an MAPPER-based manifold alignment technique by extending \cite{tuia2014semisupervised} for the semi-supervised fusion of HS and polarimetric SAR images. Besides, some follow-up researches \cite{gu2015novel,luo2017fusion,chen2017multi,liu2019stfnet,hong2020x} have been successively proposed by the attempts to enhance the capability of information blending between multi-modalities with more advanced strategies.

\subsection{Deep Models for MML}
Due to its finer and richer characterization of the scene, DL techniques \cite{hong2020x} have made great progress on multimodal image analysis and understanding. In recent years, researchers have sought to explore possibilities of using MDL and developing its variants for classifying multimodal RS images more effectively. These models can be roughly categorized into two groups.

One is the common \textit{pixel-level} multimodal classification network. Typically, Ghamisi \textit{et al.} \cite{ghamisi2016hyperspectral} extracted the EPs from HS and LiDAR data and fused them on the deep feature space generated by deep CNNs. Further, Chen \textit{et al.} \cite{chen2017deep} designed a end-to-end deep fusion network, which consists of two CNNs for feature extraction and one DNN for feature fusion. In \cite{xu2017multisource}, authors put forward to use the two-branch CNNs with cascade blocks for automatic feature extraction and fusion of multisource RS data. A general DL-based framework is developed in \cite{benedetti2018m} for the fusion of multitemporal and multimodal satellite data. 
The other MDL-based family aims to assign a semantic category to each pixel in the \textit{object-level} fashion, also known as semantic segmentation (SS). A representative model proposed by Audebert \textit{et al.} \cite{audebert2016semantic} is to segment multimodal EO data -- high-resolution RGB and DSM images -- using a multi-scaled network design. The same researchers further extended their model with two kinds of fusion strategies: early fusion and late fusion \cite{audebert2018beyond}. Another interesting work \cite{volpi2018deep} derives from a geographically-regularized deep multi-task networks for SS in aerial images. Srivastava \textit{et al.} \cite{srivastava2019understanding} provided an MDL's solution to enhance the understanding of urban land use from both overhead and ground images. Lately, the winners in 2018 IEEE Data Fusion Contest (DFC) reported their SS results via a fused fully convolutional network (FCN) conducted on MS-LiDAR and HS data \cite{xu2019advanced}. It should be noted, however, that segmentation networks usually reply on abundant labeled images and high-resolution data sources. This not only poses a great challenge in saving time and cost, but also is relatively difficult to classify accurately with small samples. \textit{Thus, this paper mainly bends our efforts for pixel-level classification tasks of multimodal RS images.}

\begin{figure*}[!t]
	  \centering
		\subfigure[early fusion]{
			\includegraphics[height=0.125\textheight]{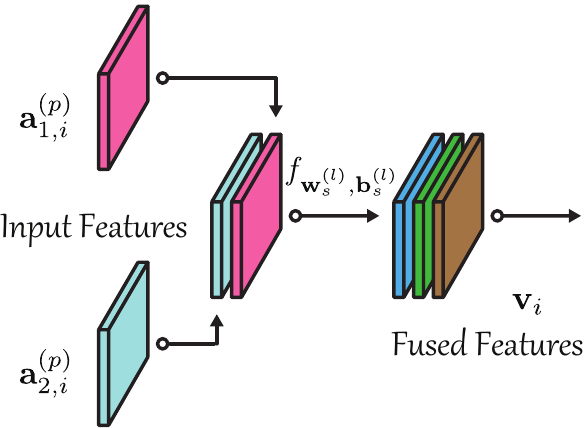}
		}
		\subfigure[middle fusion]{
			\includegraphics[height=0.125\textheight]{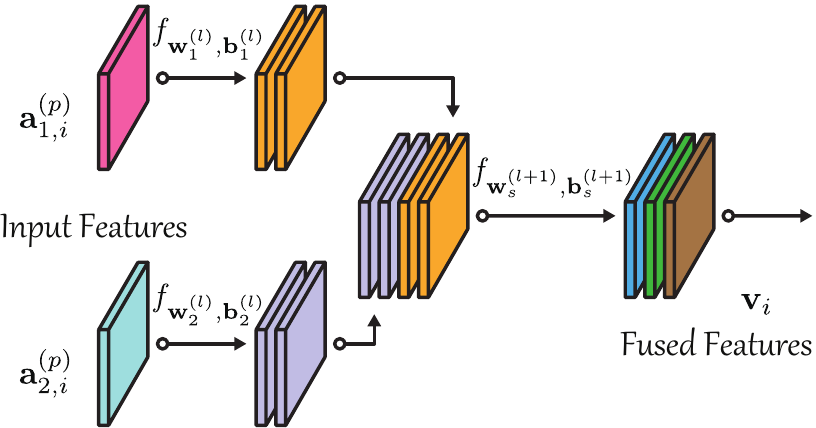}
		}
		\subfigure[late fusion]{
			\includegraphics[height=0.125\textheight]{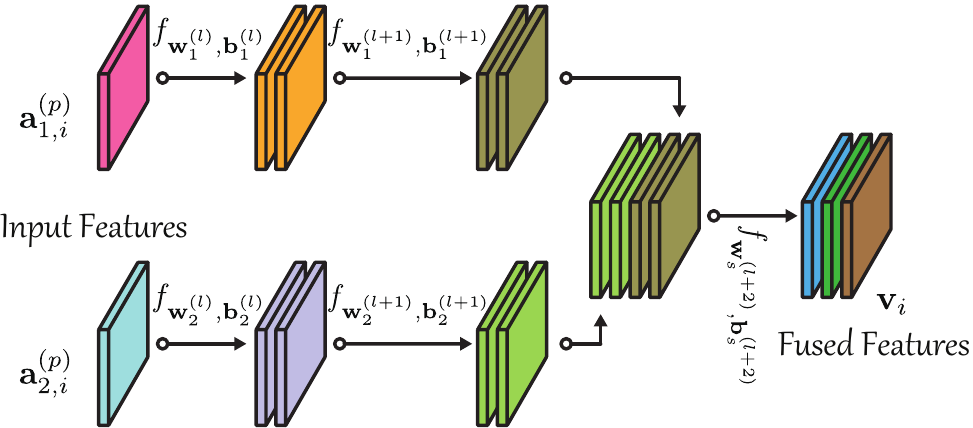}
		}
		\subfigure[encoder-decoder fusion]{
			\includegraphics[height=0.125\textheight]{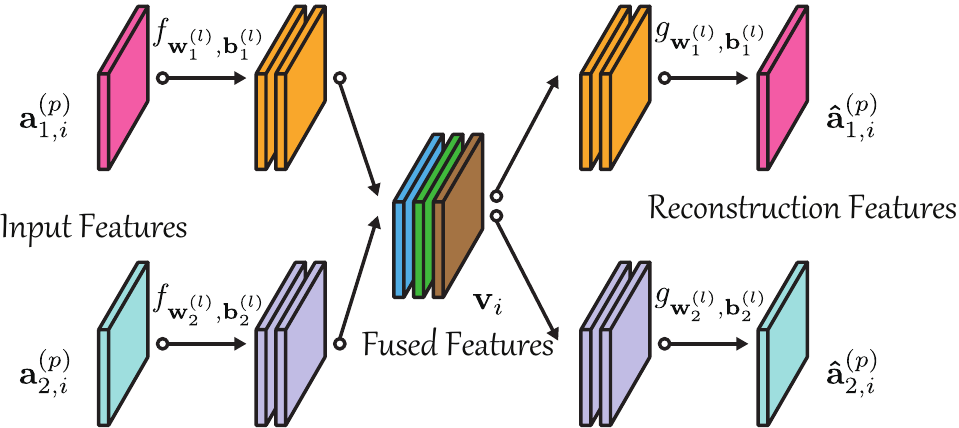}
		}\quad
	    \subfigure[cross fusion]{
			\includegraphics[height=0.131\textheight]{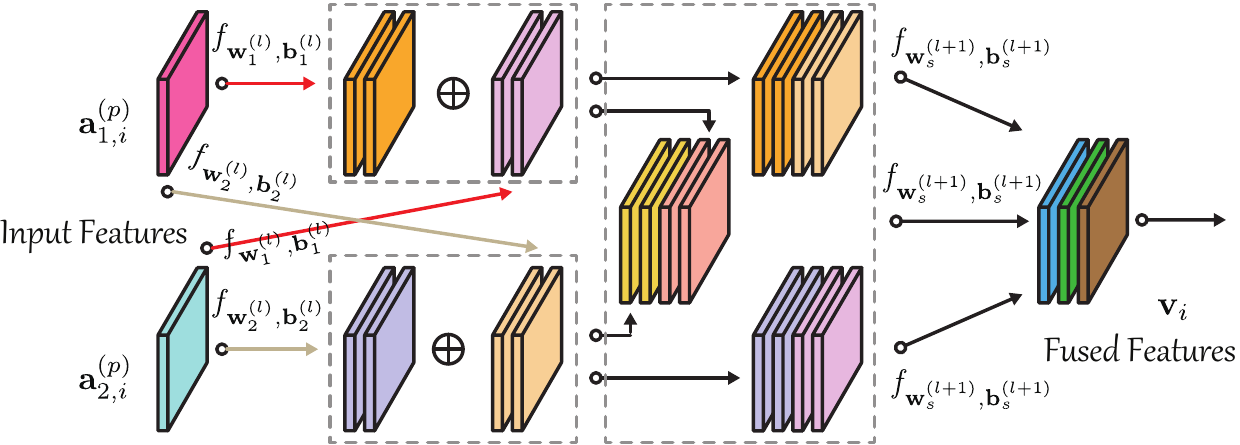}
		}
         \caption{A patch-based illustration for several plug-and-play fusion modules in the \textit{Fu-Net} of the MDL-RS framework. (a) early fusion, (b) middle fusion, (c) late fusion, (d) encoder-decoder fusion, and (e) cross fusion, where (a)-(c) are the concatenation-based fusion and (d)-(e) are the compactness-based fusion.}
\label{fig:fusion}
\end{figure*}

\subsection{CML: A Special Case of MML}
As a special family of MML, CML aims to train a model that is able to achieve a same or closer performance using either a certain modality or multiple modalities as the input  during inference process, as illustrated in Fig. \ref{fig:CML_TE}. Very recently, there has been an increasing attention on the study related to CML. Sun \textit{et al.} \cite{sun2014enhancement} made an attempt at spectrally enhancing MS imagery with partially overlapped HS data. The proposed method is a simple but feasible solution to the CML's issue. A similar work was also presented in \cite{malec2015capability} to investigate the impact of spectral enhancement on soil erosion by unmixing-based evaluation. Another stream for this topic is to directly perform feature-level learning instead of image or spectrum-level fusion. Volpi \textit{et al.} \cite{volpi2015spectral} employed the kernelized canonical correlation analysis (KCCA) to measure the dependencies between cross-sensor images for the change detection task. Hong \textit{et al.} \cite{hong2019cospace,hong2020learning} learned a common subspace from a small overlapped area of HS and MS images. The subspace can be regarded as a ``bridge'' to connect the two modalities and transfer more diverse information from one to another more effectively, particularly for larger-coverage mapping. Beyond supervision, Hong \textit{et al.} \cite{hong2019learnable} further extended their model to a semi-supervised version by learning a graph structure for alignment of labeled and unlabeled samples. We observed that data acquisition on a large scale remains challenging with an emphasis to the need of aligned multimodal sources. As compared to the case of MML, boosting the development of CML is therefore becoming more deserving in practical RS applications, e.g., large-scale classification. Yet it is relatively less investigated by RS researchers, especially in DL-guided classification tasks.

\section{Methodology}
\subsection{Method Overview}
We aim at developing a generic end-to-end multimodal deep network for RS imagery classification. The MDL-RS is shaped in the two different forms: pixel-wise and spatial-spectral architectures designed by FC-Nets and CNNs. Further, the two versions are both composed of two key modules with a focus on feature representation learning of multimodal data: \textit{Extraction Network (Ex-Net)} and \textit{Fusion Network (Fu-Net)}. Fig. \ref{fig:workflow_MDL} illustrates a general overview of the MDL-RS framework. Intuitively, the proposed MDL-RS jointly trains two subnetworks (\textit{Ex-Net} and \textit{Fu-Net}) in an end-to-end fashion.

\subsection{Extraction Network (Ex-Net)}
Our MDL-RS starts with a feature extraction network, that is \textit{Ex-Net}, which extracts hierarchical representations from different modalities. These extracted features (on the feature space) enable better information blending, particularly heterogeneous data (e.g., from different sensors) which usually fail to be fused well on the original space.

Let $\mathbf{X}_{1}\in\mathbb{R}^{d_{1}\times N}$ and $\mathbf{X}_{2}\in\mathbb{R}^{d_{2}\times N}$ be different modalities with $d_{1}$ and $d_{2}$ dimensions, respectively, by $N$ pixels, where $\mathbf{x}_{1,i}$ and $\mathbf{x}_{2,i}$ denote as an aligned $i$-th pixel-pair. The two modalities share the same label information, denoted as $\mathbf{Y} \in\mathbb{R}^{C\times N}$ with $C$ categories by $N$ pixels, which is a one-hot encoded label matrix. With these definitions, the output in the $l$-th layer of \textit{Ex-Net} can be then written as
\begin{equation}
\label{eq1}
\begin{aligned}
       \mathbf{z}^{(l)}_{s,i}=
       \begin{cases}
       h_{\mathbf{W}_{s}^{(l)}, \mathbf{b}_{s}^{(l)}}(\mathbf{x}_{s,i}), & l=1, \\
       h_{\mathbf{W}_{s}^{(l)}, \mathbf{b}_{s}^{(l)}}(\mathbf{z}_{s,i}^{(l-1)}), & l=2,...,p,
       \end{cases}
\end{aligned}
\end{equation}
where $s=0,1,2$ denotes different network streams, in particular, $s=1,2$ for different modalities and $s=0$ for the fusion stream. Here, $h(\cdot)$ is defined as the linear regression function (e.g., encoder or convolutional operation) with respect to the to-be-learned weights $\{\mathbf{W}_{s}^{(l)}\}_{l=1}^{p}$ and biases $\{\mathbf{b}_{s}^{(l)}\}_{l=1}^{p}$ of all layers ($l=1,2,...,p$) in the \textit{Ex-Net}. Inspired by the success of a batch normalization (BN) operation \cite{ioffe2015batch} that can speed up the network convergence and alleviate the problems of exploding or vanishing gradients by reducing the internal covariance shift between samples, a BN layer is then added over the output $\mathbf{z}_{s,i}^{(l)}$
\begin{equation}
\label{eq2}
\begin{aligned}
      \mathbf{z_{BN}}^{(l)}_{s,i}=\gamma_{s}\mathbf{\hat{z}}_{s,i}^{(l)}+\beta_{s},
\end{aligned}
\end{equation}
where $\mathbf{\hat{z}}_{s,i}^{(l)}$ is the $z$-score result of $\mathbf{z}_{s,i}^{(l)}$, $\gamma_{s}$ and $\beta_{s}$ denote the learnable network parameters for the $s$-th network (or modality) stream. Before importing the $\mathbf{z_{BN}}^{(l)}_{s,i}$ into the next block\footnote{We define the sequence of encoder (or convolution) operation, BN, and nonlinear activation as a block in networks.}, we have the following output ($\mathbf{a}_{s,i}^{(l)}$) behind an nonlinear activation function
\begin{equation}
\label{eq3}
\begin{aligned}
      \mathbf{a}_{s,i}^{(l)}=u(\mathbf{z_{BN}}^{(l)}_{s,i}).
\end{aligned}
\end{equation}
Here, $u(\cdot)$ is defined as the nonlinear activation function, which is performed by ${\rm ReLU}$, i.e.,
\begin{equation}
\label{eq4}
\begin{aligned}
      u(\cdot)={\rm max}(\mathbf{0}, \; \cdot).
\end{aligned}
\end{equation}

\begin{figure*}[!t]
	  \centering
		\subfigure[concatenation-based fusion]{
			\includegraphics[width=0.315\textwidth]{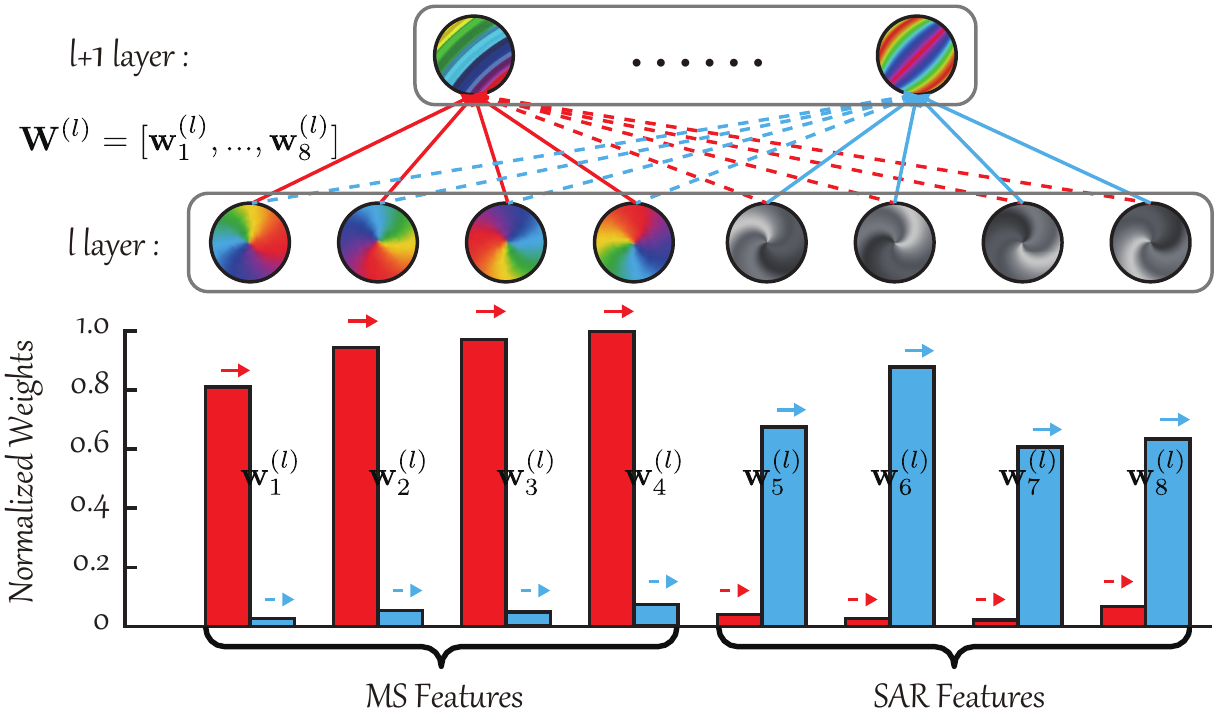}
            \label{fig:concat}
		}
		\subfigure[encoder-decoder fusion]{
			\includegraphics[width=0.315\textwidth]{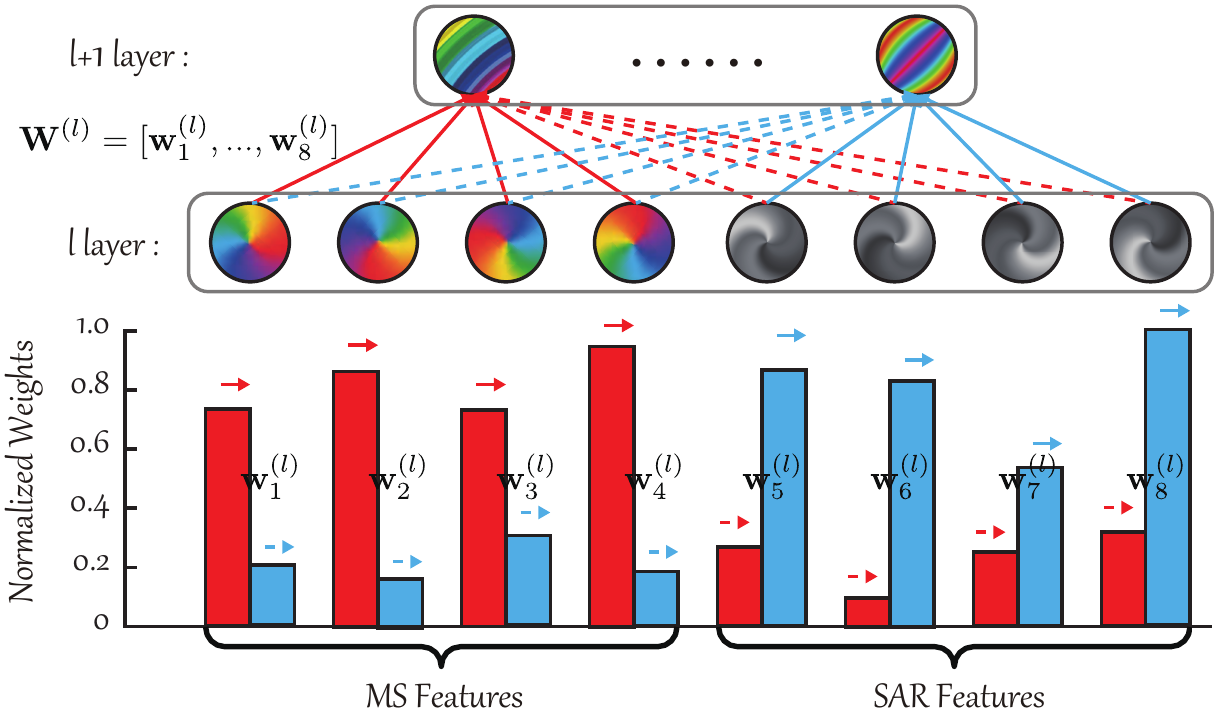}
            \label{fig:en_de}
		}
		\subfigure[cross fusion]{
			\includegraphics[width=0.315\textwidth]{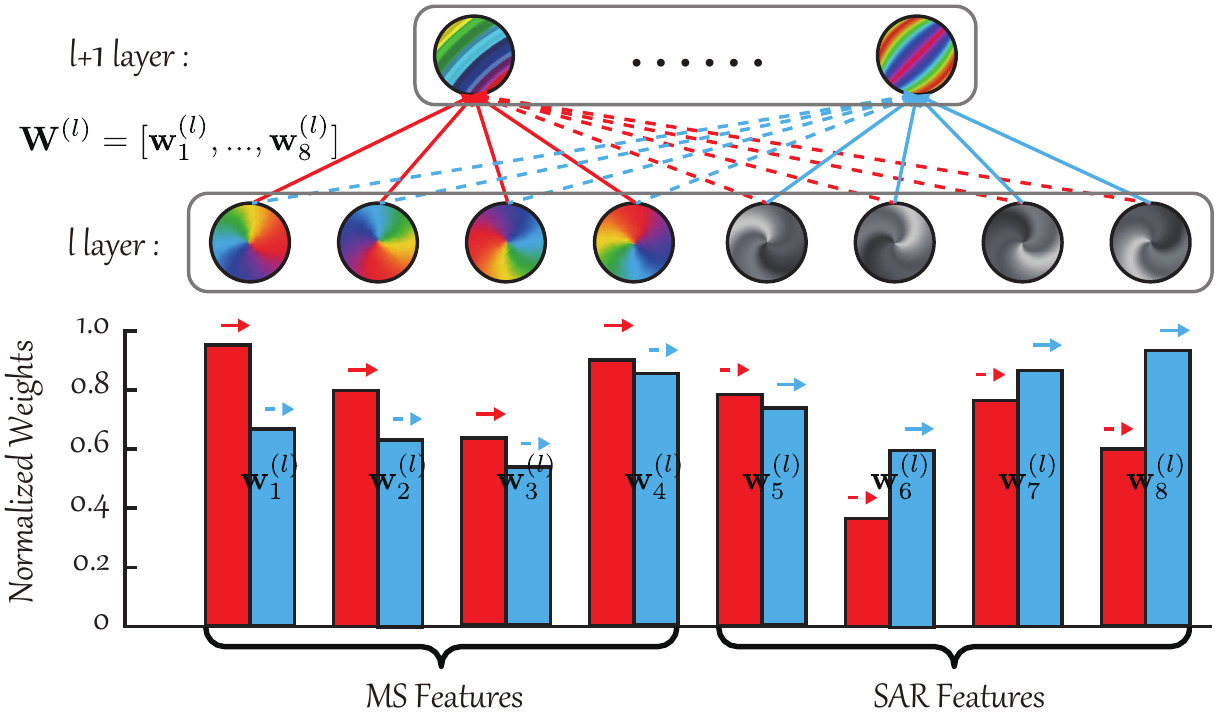}
            \label{fig:cross}
		}
         \caption{A visualization example of neurons activation across the heterogeneous MS and SAR data in the fusion layer of the \textit{Fu-Net} by comparing different fusion modules used in our MDL-RS framework. They are (a) concatenation-based (middle) fusion, (b) encoder-decoder fusion, and (c) cross fusion, in which both (b) and (c) belong to the compactness-based fusion. In detail, colorful and gray solid circles denote the feature representations of MS and SAR data in the $l$-th layer, respectively, The red and blue colors, e.g., lines and histograms, represent the network weights used to obtain the next-layer feature representations, where the dashed lines mean the weight contributions from another modality.}
\label{fig:Compact_CML}
\end{figure*}

\subsection{Fusion Network (Fu-Net)}
Once the input modalities $\mathbf{X}_{1}$ and $\mathbf{X}_{2}$ pass through the \textit{Ex-Net}, their encoded features, denoted as $\{\mathbf{A}_{s}=[\mathbf{a}_{s,1}^{(p)},...,\mathbf{a}_{s,N}^{(p)}]\}_{s=1}^{2}$, can be regarded as the new input and fed into the \textit{Fu-Net} in an end-to-end fashion. Using a similar block of \textit{Ex-Net}, e.g., Eqs. (\ref{eq1}) to (\ref{eq3}), the output of \textit{Fu-Net} can be generalized to
\begin{equation}
\label{eq5}
\begin{aligned}
        \mathbf{a}_{i}^{(l)}=f_{\mathbf{W}_{s}^{(l)}, \mathbf{b}_{s}^{(l)}}(\mathbf{a}_{s,i}^{(p)}), \;\; l=p+1,...,q,
\end{aligned}
\end{equation}
where $f(\cdot)$ denotes the nonlinear mapping function that consists of several blocks in the \textit{Fu-Net}. By investigating ``how to fuse'',  we will unfold the \textit{Fu-Net} in our MDL-RS framework to the following two groups.

\subsubsection{Concatenation-based fusion}
An intuitive fusion way in \textit{Fu-Net} is to simply stack the outputs derived from the different streams in networks. According to the requirement of ``where to fuse'', the fusion manner can be further categorized into \textit{early fusion}, \textit{middle fusion}, and \textit{late fusion} \cite{bloesch2018codeslam,feng2019deep}, as shown in Figs. \ref{fig:fusion}(a)-\ref{fig:fusion}(c). Hence, the vector representation ($\mathbf{v}_{i}$) in the $i$-th pixel corresponding to the aforementioned three fusion strategies are successively written as
\begin{equation}
\label{eq6}
\begin{aligned}
       \mathbf{v}_{i}=
       \begin{cases}
       [\mathbf{a}_{1,i}^{(l)},\;\mathbf{a}_{2,i}^{(l)}], & l=p, \\
       [f_{\mathbf{W}_{1}^{(l)},\mathbf{b}_{1}^{(l)}}(\mathbf{a}_{1,i}^{(p)}),\;f_{\mathbf{W}_{2}^{(l)},\mathbf{b}_{2}^{(l)}}(\mathbf{a}_{2,i}^{(p)})], & p < l < q,\\
       [f_{\mathbf{W}_{1}^{(l)},\mathbf{b}_{1}^{(l)}}(\mathbf{a}_{1,i}^{(p)}),\;f_{\mathbf{W}_{2}^{(l)},\mathbf{b}_{2}^{(l)}}(\mathbf{a}_{2,i}^{(p)})], & l=q,\\
       \end{cases}
\end{aligned}
\end{equation}
where $\forall l \in \mathcal{Z}$ (integer set), and ``[$\cdot$,$\cdot$]'' denotes the usual concatenation.

\subsubsection{Compactness-based fusion} \textit{Fu-Net} aims to learn better features over multiple modalities. Although the widely-used concatenation-based fusion has shown its success in feature extraction and representation, yet the capability in blending different proprieties, especially for heterogeneous data, remains limited. Alternatively, a feasible solution is to fuse the features of different modalities in a more compact way.

One representative approach presented in \cite{ngiam2011multimodal} is the \textit{En-De fusion} (see Fig. \ref{fig:fusion}(d) for details), which can be performed by minimizing the following reconstruction loss
\begin{equation}
\label{eq7}
\begin{aligned}
     \mathop{\min}_{\phi,\varphi}&\sum_{s=1}^{2}\norm{\mathbf{X}_{s}-g_{\varphi}(f_{\phi}(\mathbf{X}_{s}))}^{2}_{\F},\\
     &{\rm where} \;\; \{f_{\phi}(\mathbf{X}_{s})\}_{s=1}^{2}\rightarrow\mathbf{V}:=\{\mathbf{v}_{i}\}_{i=1}^{N}.
\end{aligned}
\end{equation}
$\norm{\cdot}_{\F}$ is the Frobenius Norm, and $f_{\phi}(\cdot)$ and $g_{\varphi}(\cdot)$ are defined as the encoder and the reconstruction-based decoder with respect to the to-be-estimated variable sets $\phi:=\{\mathbf{W}_{s}^{(l)},\mathbf{b}_{s}^{(l)}\}_{l=1}^{p}$ and $\varphi:=\{\mathbf{\tilde{W}}_{s}^{(l)},\mathbf{\tilde{b}}_{s}^{(l)}\}_{l=1}^{p}$, respectively.

Another plug-and-play fusion module proposed in this paper is named as \textit{cross fusion}. As the name suggests, the module seeks to learn more compact feature representations across modalities by interactively updating the parameters of different subnetworks. Owing to such a setting, the network stream for one modality is capable of not only learning the specific properties from itself but also considering more diversified supplement from another stream towards a more sufficient information blending. Taking the $i$-th pixel as an example, the fusion representation then is
\begin{equation}
\label{eq8}
\begin{aligned}
     &\mathbf{a}_{1,i}^{(l)}=f_{\mathbf{W}_{1}^{(l)},\mathbf{b}_{1}^{(l)}}(\mathbf{a}_{1,i}^{(p)}) + f_{\mathbf{W}_{1}^{(l)},\mathbf{b}_{1}^{(l)}}(\mathbf{a}_{2,i}^{(p)}),\\
     &\mathbf{a}_{2,i}^{(l)}=f_{\mathbf{W}_{2}^{(l)},\mathbf{b}_{2}^{(l)}}(\mathbf{a}_{2,i}^{(p)}) + f_{\mathbf{W}_{2}^{(l)},\mathbf{b}_{2}^{(l)}}(\mathbf{a}_{1,i}^{(p)}),\\
     &\mathbf{v}_{i} = \left[
        \begin{matrix}
             f_{\mathbf{W}_{0}^{(l+1)},\mathbf{b}_{0}^{(l+1)}}(\mathbf{a}_{1,i}^{(l)}), & f_{\mathbf{W}_{0}^{(l+1)},\mathbf{b}_{0}^{(l+1)}}(\mathbf{a}_{2,i}^{(l)})\\
             f_{\mathbf{W}_{1}^{(l)},\mathbf{b}_{1}^{(l)}}(\mathbf{a}_{1,i}^{(p)}), & f_{\mathbf{W}_{2}^{(l)},\mathbf{b}_{2}^{(l)}}(\mathbf{a}_{1,i}^{(p)})\\
             f_{\mathbf{W}_{1}^{(l)},\mathbf{b}_{1}^{(l)}}(\mathbf{a}_{2,i}^{(p)}), & f_{\mathbf{W}_{2}^{(l)},\mathbf{b}_{2}^{(l)}}(\mathbf{a}_{2,i}^{(p)})\\
        \end{matrix}
     \right],
\end{aligned}
\end{equation}
where the three components (each row of the matrix) of $\mathbf{v}_{i}$ in Eq. (\ref{eq8}) share the same to-be-learned parameters. In other words, they can be also seen as three ``new'' different samples for the input of the next layer to enforce a more compact fusion. Fig. \ref{fig:fusion}(e) illustrates the interactive process in networks, where highly compact fusion via crossing weights and features enables ``better'' and ``more effective'' fusion representations. More specifically, the learned weights are used across modalities, e.g., the weights learned from modality A can be simultaneously acted on modality B and \textit{vice versa}. Then, the features after summation operation together with cross combination of features again are output as the final fusion representations of \textit{cross fusion} (please see Fig. \ref{fig:fusion}(e) and Eq. (\ref{eq8}) for more details).

\begin{table*}[!t]
\centering
\caption{General network configuration in each layer of our MDL-RS framework: pixel-wise MDL-RS with FC-Nets and spatial-spectral MDL-RS with CNNs. FC, Conv, MP, and AP are abbreviations of fully connected, convolution, max pooling, and average pooling, respectively, while $d$ and $C$ denote the dimension of input and output, respectively. Moreover, the last component in each block shows its output size.}
\resizebox{0.95\textwidth}{!}{
\begin{tabular}{l||c|cccc|ccc|c}
\toprule[1.5pt]
\multirow{2}{*}{MDL-RS} & \multirow{2}{*}{Input} & \multicolumn{4}{c|}{\textit{Ex-Net}} & \multicolumn{3}{c|}{\textit{Fu-Net}} & \multirow{2}{*}{Output}\\
\cline{3-9}
& & Block1 & Block2 & Block3 & Block4 & Block5 & Block6 & Block7 & \\
\hline
\multirow{4}{*}{FC-Nets} & \multirow{4}{*}{$d$} & FC & FC & FC & FC & FC & FC & FC & \multirow{4}{*}{$C$}\\
& & BN & BN & BN & BN & BN & BN & -- & \\
& & ReLU & ReLU & ReLU & ReLU & ReLU & ReLU & Softmax & \\
& & $16$ & $32$ & $64$ & $128$ & $128$ & $64$ & $C$ & \\
\hline
\multirow{5}{*}{CNNs} & \multirow{5}{*}{$7\times 7\times d$} & $3\times 3$ Conv & $1\times 1$ Conv & $3\times 3$ Conv & $1\times 1$ Conv & $1\times 1$ Conv & $1\times 1$ Conv & $1\times 1$ Conv & \multirow{5}{*}{$C$}\\
& & BN & BN & BN & BN & BN & BN &-- & \\
& & ReLU & ReLU & ReLU & ReLU & ReLU & ReLU & Softmax & \\
& & -- & $2\times 2$ MP & -- & $2\times 2$ MP & -- & $2\times 2$ AP & -- & \\
& & $7\times 7\times 16$ & $4\times 4\times 32$ &  $4\times 4\times 64$ & $2\times 2\times 128$ & $2\times 2\times 128$ & $1 \times 1\times 64$ & $1\times 1\times C$ & \\
\bottomrule[1.5pt]
\end{tabular}
}
\label{tab:network_configuration}
\end{table*}

\subsection{Significance of Compact Blending in CML}
Up to the present, a large amount of EO data, e.g., MS, SAR, have been freely available, thus making it possible to yield a large-scale and even global scale mapping (or classification). Despite so, the data with richer spatial information, such as HS images, are hardly acquired on a large scale, due to the costly storage and limitations of imaging techniques. In this connection, CML may be an effective solution to break the performance bottleneck of current models in classification accuracy by learning better feature representations over multiple source data during model training.

We found, however, that massive connections in the concatenation-based fusion module occur in variables from the same modality but few neurons across the modalities are activated, even if each modality passes through individual \textit{Ex-Net} before being fed into the fusion layer. As illustrated in Fig. \ref{fig:concat}, where it is obvious that the neurons from one modality are activated and those from another modality are inhibited while the next-layered representations in the networks are learned. By contrast, the \textit{encoder-decoder fusion} strategy, as shown in Fig. \ref{fig:en_de}, as a member of the compactness-based fusion, can alleviate the problem to some extent. More significantly, the newly-proposed \textit{cross fusion} module is capable of blending these heterogeneous data more sufficiently. As shown in Fig. \ref{fig:cross}, the learned weights $\mathbf{W}$ across heterogeneous modalities can be balanced effectively as the subnetworks are jointly updated by the means of the mutual constraint (or transfer) between different properties.

\begin{table}[!t]
\centering
\caption{A list of the number of training and testing samples for each class in Houston2013 datasets.}
\begin{tabular}{cccc}
\toprule[1.5pt]
No. & Class Name & Training & Testing\\
\hline \hline 1 & Healthy Grass & 198 & 1053\\
 2 & Stressed Grass & 190& 1064\\
 3 & Synthetic Grass & 192 & 505\\
 4 & Tree & 188 & 1056\\
 5 & Soil & 186 & 1056\\
 6 & Water & 182 & 143\\
 7 & Residential & 196 & 1072\\
 8 & Commercial & 191 & 1053\\
 9 & Road & 193 & 1059\\
 10 & Highway & 191 & 1036\\
 11 & Railway & 181 & 1054\\
 12 & Parking Lot1 & 192 & 1041\\
 13 & Parking Lot2 & 184 & 285\\
 14 & Tennis Court & 181 & 247\\
 15 & Running Track & 187 & 473\\
\hline \hline & Total & 2832 & 12197\\
\bottomrule[1.5pt]
\end{tabular}
\label{Table:H2013}
\end{table}

\subsection{Network Architecture for MDL-RS}
As we mentioned, the proposed MDL-RS framework aims to provide a baseline network for multimodal RS imagery classification, and many plug-and-play modules can be embedded into the networks. For this purpose, we empirically and experimentally set up a basic network architecture of the MDL-RS, including two versions: pixel-wise FC-Nets and spatial-spectral CNNs, and detail them in a layer-by-layer manner. Table \ref{tab:network_configuration} lists configuration for the layer-wise network architecture. Note that there are slight differences between different fusion modules in the basic architecture, which are detailed as below.

\begin{itemize}
    \item Our MDL-RS framework for single modalities and \textit{early fusion} before feature extraction is a single-stream network for either \textit{Ex-Net} or \textit{Fu-Net}.
    \item \textit{Middle fusion}, \textit{late fusion}, \textit{en-de fusion}, and \textit{cross fusion} in the MDL-RS framework follow a two-stream \textit{Ex-Net}.
    \item The fusion behavior happens in the input for \textit{early fusion}, the Block 5 of \textit{Fu-Net} for \textit{middle fusion}, \textit{en-de fusion}, and \textit{cross fusion}, and the Block 7 of \textit{Fu-Net} for \textit{late fusion}.
    \item Unlike \textit{middle fusion}, \textit{late fusion}, and \textit{cross fusion} that hold the same setting for each layer in \textit{Fu-Net}, \textit{en-de fusion} needs to learn additional network parameters to reconstruct the fused features generated from Block 4 of \textit{Fu-Net}. The reconstruction module consists of the similar blocks with \textit{Ex-Net} by removing BN layer and replacing ReLU with Sigmoid.
    \item Considering a patch-based input in CNNs-based architecture, we spaced the pooling layer to help Conv layer to extract spatial information more effectively, where in the Block 6 average pooling (AP) is adopted rather than max pooling (MP) to reduce the loss of spatial information. Note that the strides in Conv Layer and pooling layer are both set to 1.
\end{itemize}

\begin{table}[!t]
\centering
\caption{A list of the number of training and testing samples for each class in LCZ datasets.}
\begin{tabular}{cccc}
\toprule[1.5pt]
No. & Class Name & Training (Berlin) & Testing (Hong Kong)\\
\hline \hline 1 & Compact Mid-rise & 1534 & 179\\
 2 & Open High-rise & 577 & 673\\
 3 & Open Mid-rise & 2448 & 126\\
 4 & Open Low-rise & 4010 & 120\\
 5 & Large Low-rise & 1654 & 137\\
 6 & Dense Trees & 4960 & 1616\\
 7 & Scattered Trees & 1028 & 540\\
 8 & Bush and Scrub & 1050 & 691\\
 9 & Low Plants & 4424 & 985\\
 10 & Water & 1732 & 1603\\
\hline \hline & Total & 23417 & 7670 \\
\bottomrule[1.5pt]
\end{tabular}
\label{Table:LCZ2017}
\end{table}

\section{Experiments}

\subsection{Data Description}
In the experiments, two multimodal datasets, including HS-LiDAR and MS-SAR data, are used for performance assessment both quantitatively and qualitatively. A brief description for the two datasets is given as follows.

\subsubsection{HS-LiDAR Houston2013 data}
\begin{figure*}[!t]
	  \centering
		\subfigure{
			\includegraphics[width=1\textwidth]{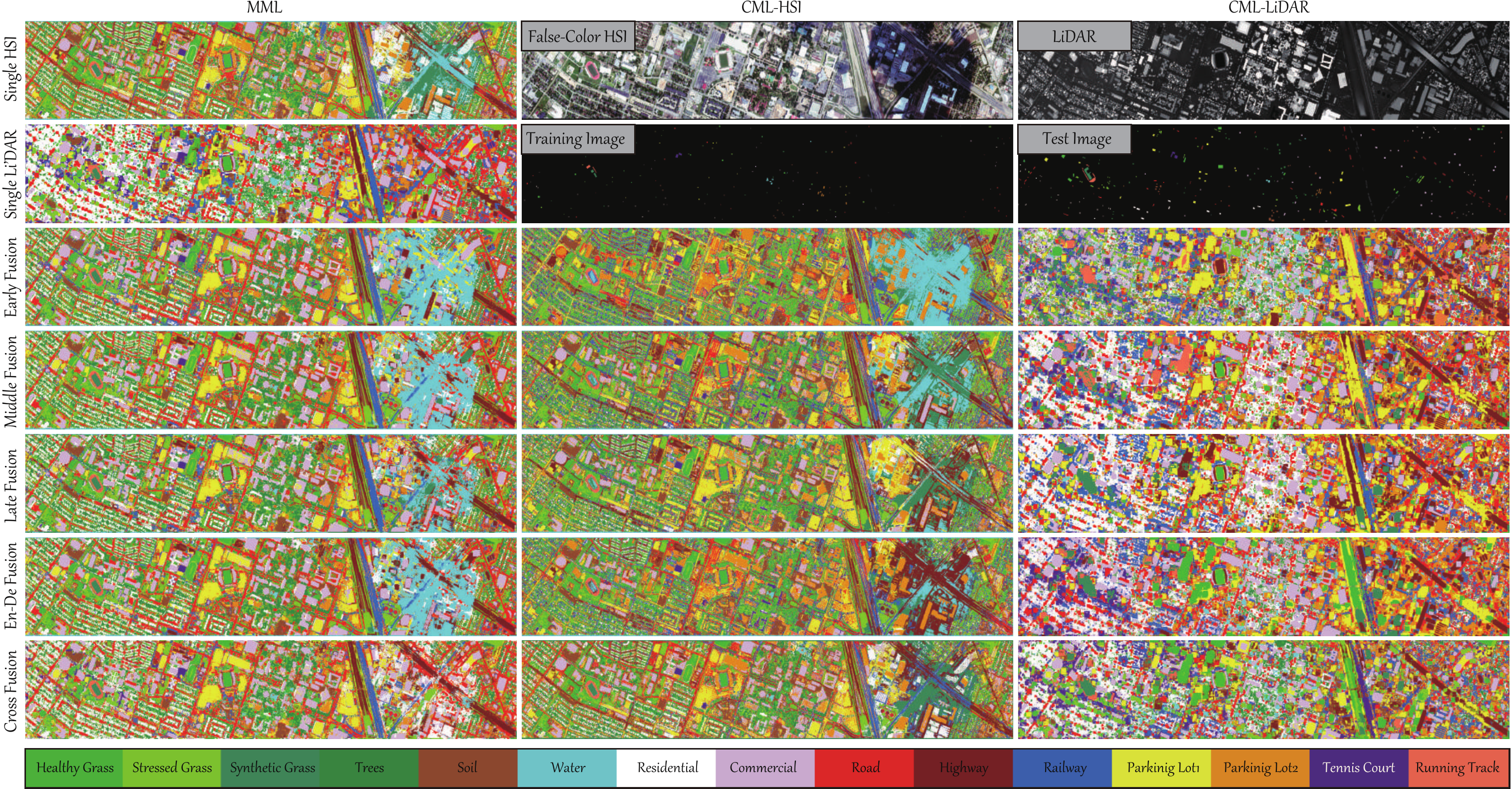}
		}
            \caption{Visualization of false-color HS and LiDAR images, the distribution of training and testing samples, and classification maps of different compared methods using FC-Nets on the HS-LiDAR Houston2013 data.}
\label{fig:CM_HH_FC}
\end{figure*}

The HS product was acquired by the ITRES CASI-1500 imaging sensor over the campus of University of Houston and its surrounding rural areas in Texas, USA, which was released for the IEEE GRSS DFC2013\footnote{http://www.grss-ieee.org/community/technical-committees/data-fusion/2013-ieee-grss-data-fusion-contest/}. The datasets consist of two data sources with $144$ bands covering the wavelength range from $364$nm to $1046$nm at a $10$nm spectral interval for HS image, and $1$ band for LiDAR data, by $349\times 1905$ pixels. Moreover, 15 LULC-related categories are investigated in the scene, whose details in terms of the class names and the size of training and testing sets are listed in Table \ref{Table:H2013}, while Fig. \ref{fig:CM_HH_FC} shows false-color images of the studied scene and the distributions of training and testing samples applied for the classification task .

\begin{table*}[!t]
\centering
\caption{Quantitative comparison of different methods using FC-Nets on the HS-LiDAR datasets. The best is shown in bold.}
\resizebox{\textwidth}{!}{
\begin{tabular}{p{23pt}<{\centering}||p{14pt}<{\centering}p{14pt}<{\centering}p{14pt}<{\centering}p{14pt}<{\centering}p{14pt}<{\centering}p{14pt}<{\centering}p{14pt}<{\centering}p{14pt}<{\centering}p{14pt}<{\centering}p{14pt}<{\centering}p{14pt}<{\centering}p{14pt}<{\centering}p{14pt}<{\centering}p{14pt}<{\centering}p{14pt}<{\centering}|p{14pt}<{\centering}p{14pt}<{\centering}p{14pt}<{\centering}}
\toprule[1.5pt]
Method & C1 & C2 & C3 & C4 & C5 & C6 & C7 & C8 & C9 & C10 & C11 & C12 & C13 & C14 & C15 & OA & AA & $\kappa$\\
\hline \hline
\multicolumn{19}{c}{\textbf{$\mathcal{MML}$}}\\
\hline \hline
HSI & 82.72 & 83.36 & \bf 100 & 92.05 & 98.20 & 95.10 & 82.84 & 48.53 & 74.88 & 52.80 & 80.74 & 84.25 & 75.79 & \bf 100 & 98.73 & 80.39 & 83.33 & 78.83\\
LiDAR & 35.71 & 57.33 & 83.56 & 72.16 & 70.08 & 71.33 & 72.29 & 85.57 & 47.88 & 60.14 & 79.70 & 47.36 & 70.53 & 85.83 & 41.23 & 63.61 & 65.38 & 60.59\\
\hline
Early & 80.44 & 80.73 & \bf 100 & 96.50 & 98.67 & 83.22 & 82.74 & 82.24 & 75.92 & 71.04 & 84.72 & 79.54 & 82.46 & \bf 100 & 98.52 & 84.88 & 86.45 & 83.66\\
Middle & 80.34 & 84.02 & \bf 100 & 92.90 & 99.53 & 95.10 & 82.46 & 82.05 & 86.21 & 75.87 & 85.48 & 81.65 & \bf 84.56 & \bf 100 & 98.52 & 86.62 & 88.58 & 85.56\\
Late & 82.81 & \bf 84.21 & \bf 100 & 93.09 & 99.05 & \bf 98.60 & 88.90 & 78.35 & 81.87 & 79.73 & 85.77 & 89.05 & 78.60 & \bf 100 & 98.73 & 87.60 & 89.06 & 86.59\\
\hline
En-De & 81.58 & 83.65 & \bf 100 & 93.09 & \bf 99.91 & 95.10 & 82.65 & 81.29 & \bf 88.29 & \bf 89.00 & 83.78 & \bf 90.39 & 82.46 & \bf 100 & 98.10 & 88.52 & 89.95 & 87.59\\
Cross & \bf 83.10 & 81.58 & \bf 100 & \bf 99.72 & 99.81 & 95.10 & \bf 90.02 & \bf 87.94 & 81.59 & 86.68 & \bf 89.37 & 85.69 & 83.16 & \bf 100 & \bf 98.73 & \bf 89.60 & \bf 90.83 & \bf 88.75\\
\hline \hline
\multicolumn{19}{c}{\textbf{$\mathcal{CML-HSI}$}}\\
\hline \hline
Early & 68.76 & 84.02 & 31.49 & 9.09 & 98.01 & 18.88 & 6.25 & 2.18 & 16.24 & 12.64 & 57.69 & 24.69 & 46.67 & 83.40 & 98.52 & 40.98 & 43.90 & 36.84\\
Middle & 81.48 & \bf 84.21 & 25.15 & 34.38 & 99.62 & 95.10 & 19.40 &45.58 & 45.99 & 38.61 & 69.17 & 60.33 & 44.56 & \bf 100 & \bf 98.73 & 59.07 & 62.82 & 56.00 \\
Late & 82.34 & \bf 84.21 & 94.65 & 78.03 & 98.30 & \bf 98.60 & 40.39 & 35.90 & 49.67 & 48.17 & 63.47 & 76.85 & 51.23 & \bf 100 & \bf 98.73 & 68.94 & 73.37 & 66.51\\
\hline
En-De & \bf 96.39 & 78.20 & 83.56 & \bf 91.29 & 96.88 & 96.50 & 58.49 & 44.63 & 55.15 & 53.86 & 69.26 & 66.76 & 64.21 & 98.38 & 95.56 & 73.26 & 76.61 & 71.10\\
Cross & 82.91 & 83.27 & \bf 99.60 & 88.73 & \bf 99.05 & 95.10 & \bf 81.16 & \bf 47.77 & \bf 76.68 & \bf 79.73 & \bf 83.02 & \bf 84.05 & \bf 74.04 & \bf 100 & 98.52 & \bf 82.53 & \bf 84.91 & \bf 81.11\\
\hline \hline
\multicolumn{19}{c}{\textbf{$\mathcal{CML-LIDAR}$}}\\
\hline \hline
Early & -- & 8.74 & 93.66 & 29.55 & 1.23 & 11.89 & 40.86 & 56.41 & 23.32 & 45.37 & 74.10 & 28.72 & \bf 82.81 & 30.77 & 0.21 & 33.20 & 35.18 & 28.20\\
Middle & 43.49 & 15.23 & -- & 12.31 & 10.70 & -- & 62.78 & 75.40 & 57.88 & 71.53 & 78.46 & 31.32 & 82.81 & 35.22 & 49.05 & 44.21 & 41.75 & 39.67\\
Late & 40.65 & 4.04 & \bf 100 & 14.49 & 0.57 & 1.4 & \bf 78.92 & 69.71 & \bf 70.63 & \bf 78.09 & \bf 79.03 & 44.09 & 78.60 & 11.34 & -- & 47.70 & 44.77 & 43.45\\
\hline
En-De & 45.77 & 42.86 & 92.48 & 69.79 & 31.06 & \bf 67.83 & 39.74 & 57.93 & 49.01 & 54.92 & 50.28 & 32.18 & 65.26 & 38.46 & 42.49 & 49.50 & 52.01 & 45.44\\
Cross & \bf 58.78 & \bf 53.20 & 75.25 & \bf 74.62 & \bf 61.08 & 60.14 & 67.54 & \bf 78.06 & 51.75 & 75.87 & 75.33 & \bf 69.55 & 77.54 & \bf 90.69 & \bf 75.05 & \bf 67.90 & \bf 69.63 & \bf 65.36\\
\bottomrule[1.5pt]
\end{tabular}}
\label{tab:HH_MML_FC}
\end{table*}

\begin{table*}[!t]
\centering
\caption{Quantitative comparison of different methods using CNNs on the HS-LiDAR datasets. The best is shown in bold.}
\resizebox{\textwidth}{!}{
\begin{tabular}{p{23pt}<{\centering}||p{14pt}<{\centering}p{14pt}<{\centering}p{14pt}<{\centering}p{14pt}<{\centering}p{14pt}<{\centering}p{14pt}<{\centering}p{14pt}<{\centering}p{14pt}<{\centering}p{14pt}<{\centering}p{14pt}<{\centering}p{14pt}<{\centering}p{14pt}<{\centering}p{14pt}<{\centering}p{14pt}<{\centering}p{14pt}<{\centering}|p{14pt}<{\centering}p{14pt}<{\centering}p{14pt}<{\centering}}
\toprule[1.5pt]
Method & C1 & C2 & C3 & C4 & C5 & C6 & C7 & C8 & C9 & C10 & C11 & C12 & C13 & C14 & C15 & OA & AA & $\kappa$\\
\hline \hline
\multicolumn{19}{c}{\textbf{$\mathcal{MML}$}}\\
\hline \hline
HSI & 83.00 & \bf 85.15 & 89.90 & 88.73 & 99.91 & 90.21 & 82.18 & 67.43 & 85.93 & 58.49 & 63.28 & 92.80 & 79.65 & 92.71 & 96.62 & 82.05 & 83.73 & 80.61\\
LiDAR & 40.17 & 52.44 & 71.49 & 82.10 & 59.94 & 60.84 & 80.97 & 84.43 & 56.47 & 65.35 & 88.52 & 56.10 & 70.18 & 65.59 & 79.28 & 67.35 & 67.59 & 64.67\\
\hline
Early & \bf 83.10 & 84.87 & 88.91 & 90.91 & 99.53 & 98.60 & 93.94 & 68.38 & 81.21 & 54.34 & 74.29 & 85.30 & 79.65 & \bf 96.76 & 98.52 & 83.07 & 85.22 & 81.65\\
Middle & \bf 83.10 & 85.06 & 99.60 & 91.57 & 98.86 & \bf 100 &96.64 & 88.13 & 85.93 & 74.42 & 84.54 & 95.39 & 87.37 & 95.14 & \bf 100 & 89.55 & 91.05 & 88.71\\
Late & 83.00 & 84.02 & 99.80 & 91.95 & 99.62 & 97.20 & 95.15 & 83.29 & \bf 88.10 & 69.69 & 80.74 & 89.82 & 88.77 & 94.33 & \bf 100 & 87.98 & 89.70 & 87.02\\
\hline
En-De & \bf 83.10 & 85.06 & \bf 100 & 92.61 & 99.72 & 95.80 & 95.34 & \bf 92.31 & 86.02 & 79.05 & 86.05 & 97.21 & \bf 92.28 & 93.93 & \bf 100 & 90.71 & 91.90 & 89.96\\
Cross & \bf 83.10 & 84.68 & 99.60 & \bf 92.80 & \bf 99.91 & \bf 100 & \bf 98.51 & 88.89 & 82.06 & \bf 91.41 & \bf 91.94 & \bf 99.14 & 85.61 & 95.95 & \bf 100 & \bf 91.99 & \bf 92.91 & \bf 91.33\\
\hline \hline
\multicolumn{19}{c}{\textbf{$\mathcal{CML-HSI}$}}\\
\hline \hline
Early & 82.91 & 81.95 & -- & -- & 98.39 & 46.85 & 2.24 & 30.77 & 61.00 & 46.43 & 0.47 & 28.63 & \bf 92.98 & 85.02 & 91.12 & 45.38 & 49.92 & 41.39\\
Middle & 82.62 & \bf 85.15 & \bf 94.46 & 80.40 & 99.91 & 95.80 & 32.84 & 40.93 & 75.64 & 24.61 & 58.06 & 72.24 & 92.63 & 89.88 & 96.41 & 69.19 & 74.77 & 66.91\\
Late & 81.77 & \bf 85.15 & 87.33 & 81.82 & 99.53 & 95.10 & 50.09 & 52.80 & 79.98 & 25.48 & 56.36 & 84.05 & 91.23 & 87.45 & 94.93 & 72.62 & 76.87 & 70.57\\
\hline
En-De & \bf 98.29 & 76.22 & 80.40 & \bf 91.57 & 98.77 & 83.92 & 75.47 & 67.24 & \bf 82.63 & \bf 58.78 & 46.87 & 83.19 & 89.82 & 87.85 & 94.71 & 79.23 & 81.05 & 77.48\\
Cross & 83.00 & 84.68 & 93.27 & 90.53 & \bf 100 & \bf 97.20 & \bf 87.87 & \bf 71.60 & 81.49 & 56.37 & \bf 75.52 & \bf 92.70 & 88.42 & \bf 95.95 & \bf 97.67 & \bf 84.05 & \bf 86.42 & \bf 82.73\\
\hline \hline
\multicolumn{19}{c}{\textbf{$\mathcal{CML-LIDAR}$}}\\
\hline \hline
Early & 0.66 & 21.24 & \bf 100 & 5.97 & 0.09 & 72.03 & \bf 97.57 & 84.05 & 48.73 & 8.78 & 25.33 & 10.57 & 1.75 & 0.4 & -- & 31.37 & 31.81 & 25.76\\
Middle & 30.96 & 0.85 & \bf 100 & 25.66 & 14.58 & 61.54 & 70.62 & 84.90 & 66.01 & \bf 78.96 & \bf 86.81 & 28.24 & 66.32 & 38.46 & 2.75 & 49.41 & 50.44 & 45.46\\
Late & 54.51 & -- & \bf 100 & 16.29 & 1.61 & 63.64 & 53.08 & \bf 94.30 & \bf 72.52 & 77.90 & 79.41 & 37.85 & 65.26 & 35.22 & 1.69 & 49.26 & 50.22 & 45.12\\
\hline
En-De & \bf 55.56 & 55.26 & 88.12 & 76.61 & 46.40 & 75.52 & 65.21 & 77.97 & 55.43 & 74.13 & 80.27 & 36.79 & 37.19 & \bf 74.90 & 75.26 & 63.75 & 64.97 & 60.81\\
Cross & 52.14 & \bf 69.27 & 89.50 & \bf 81.34 & \bf 67.61 & \bf 79.02 & 79.85 & 84.90 & 56.85 &65.64 & 85.29 & \bf 52.16 & \bf 77.19 & 68.02 & \bf 79.49 & \bf 71.04 & \bf 72.55 & \bf 68.63\\
\bottomrule[1.5pt]
\end{tabular}}
\label{tab:HH_MML_CNN}
\end{table*}

\subsubsection{MS-SAR LCZ data}
The LCZ datasets are collected from Sentinel-2 and Sentinel-1 satellites, where the former acquires the MS data with 10 spectral bands and the latter is able to generate the dual-polarimetric SAR data organized as a commonly-used PolSAR covariance matrix (four components) \cite{yamaguchi2005four}. To avoid the information leak in evaluating the classification performance of the models, we thoroughly separate the training and testing sets in the LCZ datasets by training the networks on the area of \textit{Berlin} and inferring the models on \textit{Hong Kong} and its surroundings. Please note that the labeled ground truth for the two cities and the Sentinel-2 MS data are available from the IEEE GRSS DFC2017\footnote{http://www.grss-ieee.org/2017-ieee-grss-data-fusion-contest/}, as detailed in Table \ref{Table:LCZ2017} and visualized in Fig. \ref{fig:CM_LCZ_FC}.

\subsection{Experimental Setup}
\subsubsection{Implementation details}
The proposed networks are implemented on the Tensorflow platform. These models are trained on the training set, and the hyper-parameters are determined using a grid search on the validation set. More specifically, ten replications are performed to randomly separate the original training set into the new training set and validation set with the percentage of 8:2 for the final network's hyper-parameters. In the training phase, we adopt the Adam optimizer with the ``exponential'' learning rate policy. The current learning rate can be updated by multiplying the base one with $(1-\frac{iter}{maxIter})^{power}$ at intervals of 30 epochs, where the initialized learning rate and power are set to 0.001 and 0.5, respectively. We initialize the subnetworks for each modality with He initialization \cite{he2015delving}. Due to the randomness in initialization, the averaged results will be reported out of ten runs. Moreover, the momentum is parameterized by 0.9, and the training batch is set to 64 and 256 in the first and second datasets, respectively. To facilitate network training and reduce overfitting, we also employ the $\ell_2$-norm regularization on weights to avoid overfitting problems. The networks would stop training when the validation loss fails to decrease.

Note that there is only one band for the LiDAR image in the HS-LiDAR data. To fully exploit the spatial information and facilitate the network learning of the MDL-RS with FC-Nets, the attribute profiles (APs) in \cite{hong2020invariant} are extracted from the LiDAR image, resulting in 21-band profiles.

Furthermore, to evaluate the models' performance more effectively, we train the networks using multi-modalities and not only infer the models in the \textbf{MML's} case with the multimodal input but also infer the models in the \textbf{CML's} issue by zeroing one of modalities (take bi-modality as an example).

\subsubsection{Evaluation metric}
Pixel-level RS image classification is explored as a potential target for evaluating the performance of the proposed MDL-RS framework. More specifically, three commonly-used indices -- \textit{Overall Accuracy (OA)}, \textit{Average Accuracy (AA)}, and \textit{Kappa Coefficient ($\kappa$)} -- are calculated to quantify classification performance. They can be formulated by using the following equations.
\begin{equation}
\label{eq9}
\begin{aligned}
      OA = \frac{N_{c}}{N_{a}},
\end{aligned}
\end{equation}
\begin{equation}
\label{eq10}
\begin{aligned}
      AA = \frac{1}{C}\sum_{i=1}^{C}\frac{N_{c}^{i}}{N_{a}^{i}},
\end{aligned}
\end{equation}
and
\begin{equation}
\label{eq11}
\begin{aligned}
      \kappa = \frac{OA-P_{e}}{1-P_{e}},
\end{aligned}
\end{equation}
where $N_{c}$ and $N_{a}$ denote the number of samples classified correctly and the number of total samples, respectively, while $N_{c}^{i}$ and $N_{a}^{i}$ correspond to the $N_{c}$ and $N_{a}$ of each class, respectively. $P_{e}$ in $\kappa$ is defined as the hypothetical probability of chance agreement \cite{cohen1960coefficient}, which can be computed by
\begin{equation}
\label{eq12}
\begin{aligned}
      P_{e} = \frac{N_{r}^{1}\times N_{p}^{1} + \dots N_{r}^{i}\times N_{p}^{i} + \dots + N_{r}^{C}\times N_{p}^{C}}{N_{a}\times N_{a}},
\end{aligned}
\end{equation}
where $N_{r}^{i}$ and $N_{p}^{i}$ denote the number of real samples for each class and the number of predicted samples for each class, respectively.

\subsubsection{Comparison with state-of-the-art baselines}
Several state-of-the-art baselines in terms of different fusion strategies are selected for comparison, including
concatenation-based fusion: \textit{early fusion}, \textit{middle fusion}, and \textit{late fusion}, and compactness-based fusion: \textit{en-de fusion} and \textit{cross fusion}, as well as single modalities. These models are also performed by using both FC-Nets and CNNs frameworks. It is worth noting, however, that the patch centered by a pixel is usually used as the input of CNNs in RS image classification. For this reason, we need to extend the original image by the ``replicate'' operation, that is, copying the pixels within the image to that out of the original image boundary, to solve the problem of the boundaries of the multimodal RS data in the CNNs-related experiments.

\begin{figure*}[!t]
	  \centering
		\subfigure{
			\includegraphics[width=1\textwidth]{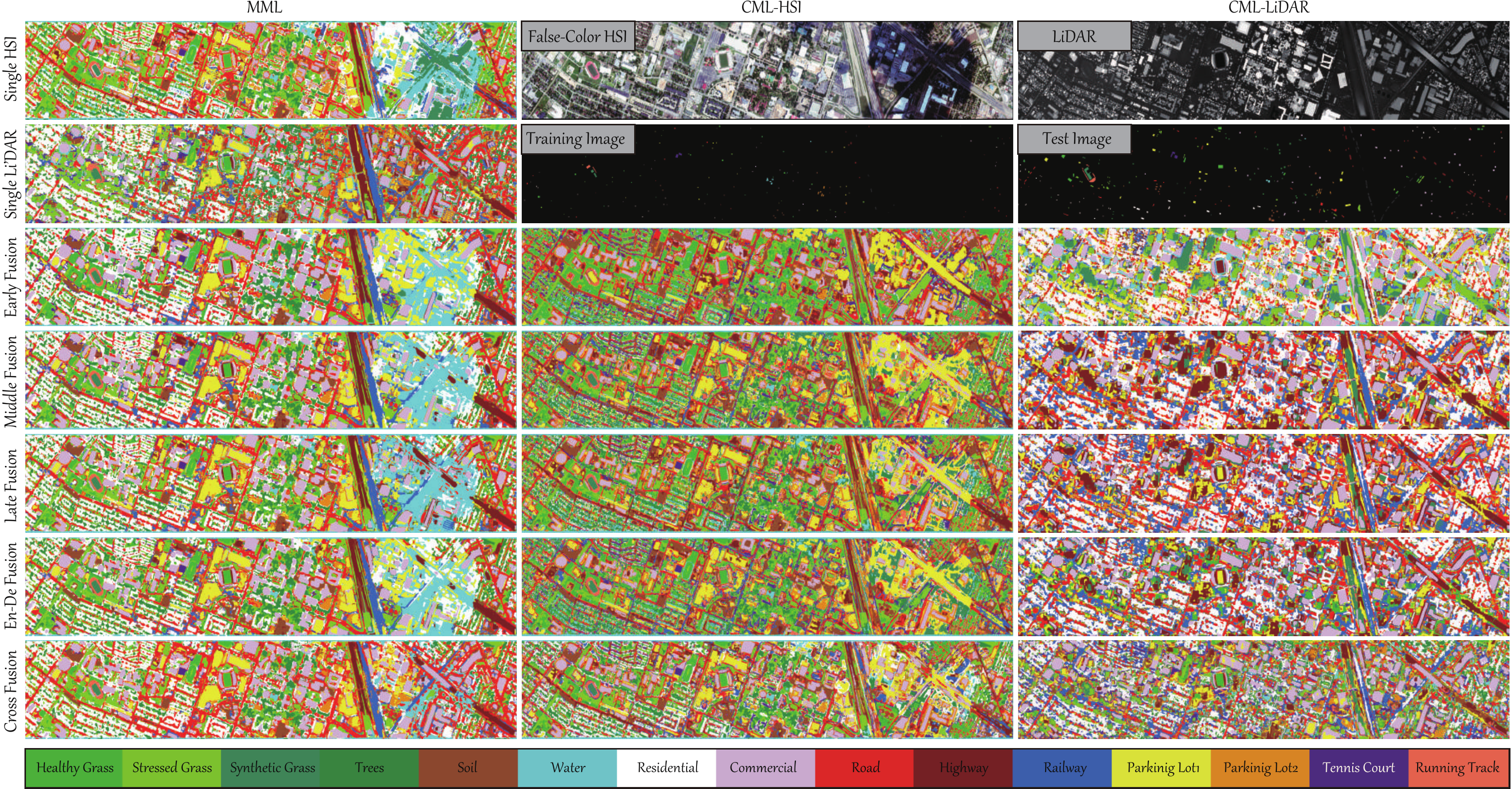}
		}
        \caption{Visualization of false-color HS and LiDAR images, the distribution of training and testing samples, and classification maps of different compared methods using CNNs on the HS-LiDAR Houston2013 data.}
\label{fig:CM_HH_CNN}
\end{figure*}

\begin{table*}[!t]
\centering
\caption{Quantitative comparison of different methods using FC-Nets on the MS-SAR datasets. The best is shown in bold.}
\resizebox{0.75\textwidth}{!}{
\begin{tabular}{p{23pt}<{\centering}||p{14pt}<{\centering}p{14pt}<{\centering}p{14pt}<{\centering}p{14pt}<{\centering}p{14pt}<{\centering}p{14pt}<{\centering}p{14pt}<{\centering}p{14pt}<{\centering}p{14pt}<{\centering}p{14pt}<{\centering}|p{14pt}<{\centering}p{14pt}<{\centering}p{14pt}<{\centering}}
\toprule[1.5pt]
Method & C1 & C2 & C3 & C4 & C5 & C6 & C7 & C8 & C9 & C10 & OA & AA & $\kappa$\\
\hline \hline
\multicolumn{14}{c}{\textbf{$\mathcal{MML}$}}\\
\hline \hline
MSI & 3.91 & \bf 5.2 & 21.43 & 2.5 & 74.45 & 89.23 & \bf 62.96 & 14.47 & 7.72 & 42.22 & 42.13 & 32.41 & 33.00\\
SAR & 6.15 & -- & 2.38 & 0.83 & 2.92 & -- & -- & -- & -- & \bf 99.62 & 34.05 & 11.19 & 0.98\\
\hline
Early & \bf 8.94 & -- & 17.46 & 10.00 & 78.10 & 94.37 & 53.52 & 7.24 & 14.11 & 51.71 & 45.71 & 33.54 & 36.59\\
Middle & 9.5 & -- & 17.46 & \bf 19.17 & \bf 78.83 & \bf 96.53 & 30.56 & \bf 25.04 & 8.73 & 53.55 & 46.26 & 33.94 & 36.73\\
Late & 0.56 & 0.15 & 6.35 & -- & 78.10 & 10.15 & 53.70 & 16.06 & 54.11 & 90.66 & 46.61 & 30.99 & 36.03\\
\hline
En-De & 3.91 & 0.59 & 6.35 & -- & 74.45 & 40.90 & 6.85 & -- & \bf 58.07 & 98.23 & 51.47 & 28.94 & 40.11 \\
Cross & 7.82 & -- & \bf 54.76 & 0.83 & 72.26 & 59.90 & 24.63 & 18.38 & 30.96 & 94.89 & \bf 54.58 & \bf 36.44 & \bf 43.97\\
\hline \hline
\multicolumn{14}{c}{\textbf{$\mathcal{CML-MSI}$}}\\
\hline \hline
Early & \bf 96.09 & \bf 0.45 & \bf 50.79 & -- & 18.25 & 46.53 & \bf 65.74 & 0.14 & 5.99 & -- & 18.66 & 28.40 & 13.13\\
Middle & -- & -- & -- & -- & -- & -- & -- & -- & 72.79 & 97.89 & 42.57 & 17.07 & 24.51\\
Late & -- & -- & -- & -- & \bf 82.48 & -- & 19.07 & 4.92 & 80.91 & 84.86 & 42.45 & 27.23 & 28.68\\
\hline
En-De & 1.12 & -- & 0.79 & -- & 11.68 & -- & -- & -- & \bf 90.96 & \bf 98.69 & 45.42 & 20.32 & 29.59\\
Cross & 10.61 & -- & 9.52 & \bf 15.00 & 76.64 & \bf 98.39 & 27.22 & \bf 8.83 & 27.92 & 63.00 & \bf 50.42 & \bf 33.71 & \bf 40.49\\
\hline \hline
\multicolumn{14}{c}{\textbf{$\mathcal{CML-SAR}$}}\\
\hline \hline
Early & -- & -- & -- & -- & -- & -- & -- & -- & -- & \bf 100 & 33.94 & 10.00 & -- \\
Middle & -- & -- & -- & -- & -- & -- & -- & -- & -- & 99.88 & 33.90 & 9.99 & -- \\
Late & -- & -- & -- & -- & -- & -- & -- & -- & -- & \bf 100 & 33.94 & 10.00 & -- \\
\hline
En-De & -- & -- & -- & -- & \bf 97.08 & 26.86 & -- & \bf 0.29 &
\bf 67.21 & 65.31 & 38.21 & \bf 25.67 & \bf 27.49\\
Cross & \bf 47.49 & -- & \bf 7.14 & \bf 19.17 & 5.84 & \bf 31.37 & -- & -- & 13.3 & 98.27 & \bf 43.30 & 22.26 & 23.47\\
\bottomrule[1.5pt]
\end{tabular}}
\label{tab:LCZ_MML_FC}
\end{table*}

\begin{table*}[!t]
\centering
\caption{Quantitative comparison of different methods using CNNs on the MS-SAR datasets. The best is shown in bold.}
\resizebox{0.75\textwidth}{!}{
\begin{tabular}{p{23pt}<{\centering}||p{14pt}<{\centering}p{14pt}<{\centering}p{14pt}<{\centering}p{14pt}<{\centering}p{14pt}<{\centering}p{14pt}<{\centering}p{14pt}<{\centering}p{14pt}<{\centering}p{14pt}<{\centering}p{14pt}<{\centering}|p{14pt}<{\centering}p{14pt}<{\centering}p{14pt}<{\centering}}
\toprule[1.5pt]
Method & C1 & C2 & C3 & C4 & C5 & C6 & C7 & C8 & C9 & C10 & OA & AA & $\kappa$\\
\hline \hline
\multicolumn{14}{c}{\textbf{$\mathcal{MML}$}}\\
\hline \hline
MSI & 1.12 & -- & 17.46 & 9.17 & \bf 91.97 & 79.89 & 72.96 & 8.54 & 11.78 & 55.28 & 45.11 & 34.82 & 36.13\\
SAR & \bf 92.18 & -- & \bf 38.10 & \bf 39.17 & 9.49 & 22.09 & 2.41 & 0.14 & 21.52 & 85.67 & 40.23 & 31.08 & 29.12\\
\hline
Early & 10.61 & 2.23 & 15.08 & 0.83 & 83.94 & 73.14 & \bf 73.33 & 13.02 & 3.25 & 79.18 & 51.24 & 35.46 & 41.65\\
Middle & 7.26 & 3.86 & 2.38 & 4.17 & 75.91 & 63.92 & 66.48 & 9.99 & 23.25 & \bf 97.04 & 56.94 & 35.43 & 46.96\\
Late & 35.75 & \bf 9.96 & -- & -- & 85.40 & \bf 98.76 & 26.85 & \bf 17.22 & 10.96 & 75.61 & 54.55 & 36.05 & 44.38\\
\hline
En-De & 24.02 & 0.59 & 26.19 & 2.5 & 72.26 & 96.16 & 43.15 & 13.31 & 10.76 & 92.28 & 59.57 & 38.12 & 49.68\\
Cross & 59.22 & 0.3 & 16.67 & 4.17 & 83.94 & 81.87 & 42.22 & 0.14 & \bf 68.02 & 91.82 & \bf 63.38 & \bf 44.84 & \bf 54.48\\
\hline \hline
\multicolumn{14}{c}{\textbf{$\mathcal{CML-MSI}$}}\\
\hline \hline
Early & 51.96 & 0.15 & 0.79 & -- & 35.04 & -- & 46.11 & 1.45 & \bf 48.22 & 64.81 & 33.43 & 24.85 & 21.92\\
Middle & \bf 59.78 & -- & \bf 27.78 & 5.00 & 83.21 & \bf 95.98 & 39.07 & 3.91 & 4.06 & 62.50 & 48.47 & 38.13 & 38.99\\
Late & 17.88 & \bf 3.27 & 22.22 & 7.50 & \bf 86.86 & 53.16 & \bf 76.30 & \bf 16.79 & 5.89 & 68.57 & 44.85 & 35.84 & 35.56\\
\hline
En-De & 3.35 & -- & 23.02 & -- & 81.02 & 92.02 & 55.37 & 11.58 & 20.61 & 77.06 & 55.03 & 36.40 & 45.47\\
Cross & 0.56 & 1.93 & 5.56 & \bf 19.17 & \bf 86.86 & 93.50 & 50.19 & 5.93 & 24.87 & \bf 91.39 & \bf 60.10 & \bf 38.00 & \bf 50.45\\
\hline \hline
\multicolumn{14}{c}{\textbf{$\mathcal{CML-SAR}$}}\\
\hline \hline
Early & -- & -- & -- & -- & -- & -- & -- & -- & -- & \bf 100 & 33.94 & 10.00 & -- \\
Middle & -- & -- & -- & -- & -- & -- & -- & -- & -- & \bf 100 & 33.94 & 10.00 & -- \\
Late & -- & -- & -- & -- & -- & -- & -- & -- & -- & \bf 100 & 33.94 & 10.00 & -- \\
\hline
En-De & 47.49 & -- & \bf 33.33 & 28.33 & \bf 1.46 & 42.08 & -- & -- & \bf 69.04 & 89.67 & 50.29 & 31.14 & 39.09\\
Cross & \bf 93.30 & -- & 7.14 & \bf 29.17 & -- & \bf 66.83 & -- & -- & 46.29 & 89.40 & \bf 53.12 & \bf 33.21 & \bf 41.67\\
\bottomrule[1.5pt]
\end{tabular}}
\label{tab:LCZ_MML_CNN}
\end{table*}

\subsection{Result and Analysis on Houston Data}
\subsubsection{Quantitative comparison}
Table \ref{tab:HH_MML_FC} lists the quantitative performance comparison in terms of \textit{OA}, \textit{AA}, and $\kappa$ as well as the accuracy for each category using a FC-based feature extractor (see FC-Nets) in three different experimental setting, e.g., MML, CML-HSI, and CML-LiDAR.

Characterized by rich spectral information, single HSI performs better than single LiDAR (over 15\% \textit{OA}), even though APs are pre-extracted from the LiDAR image before feeding into the networks. Limited by feature diversity, the single modalities yield relatively poor performance compared to those with multimodal input in MML. Moreover, the classification performance of compactness-based approaches is generally superior to that of concatenation-based ones, bringing increments of at least 1\% \textit{OA}, \textit{AA}, and $\kappa$. In details, \textit{late fusion} and \textit{middle fusion} are more effective than \textit{early fusion}, while \textit{cross fusion} outperforms others, achieving best classification results.

Regarding the CML's case, due to missing one modality in the inference process, those concatenation-based fusion approaches basically fail to work well, particularly \textit{early fusion} whose classification performance decreases dramatically to 28.13\% OA in CML-HSI and 12.76\% OA in CML-LiDAR. Although other two strategies seem to be acceptable to some extent, yet their results are even lower than those using single modalities. This might indicate that the above methods are not feasible to the CML's issue in practical applications. By contrast, the compactness-based \textit{cross fusion} overcomes other competitors either in MML's or in CML's task. More significantly, the resulting model trained by \textit{cross fusion} is capable of transferring the information from one modality to another one more effectively, yielding a higher classification accuracy than that using single modalities. In addition, the compactness-based fusion networks also behaves superiorly compared to the concatenation-based models from the perspective of per-class performance.

Furthermore, Table \ref{tab:HH_MML_CNN} shows the corresponding results obtained by CNNs. Overall, these models with the CNNs-based architecture hold a higher-level classification performance compared to those with FC-Nets (\textit{cf.} Table \ref{tab:HH_MML_FC}). The classification accuracies for all compared algorithms increase by 2\%$\sim$3\% in terms of three main indices as a whole. The benefits of the CNNs-based network design are, on the one hand, to extract the semantically meaningful information from locally neighboring pixels; and, on the other hand, able to perform the information blending more sufficiently in a spatial-spectral fashion.

\begin{figure*}[!t]
	  \centering
		\subfigure{
			\includegraphics[width=0.95\textwidth]{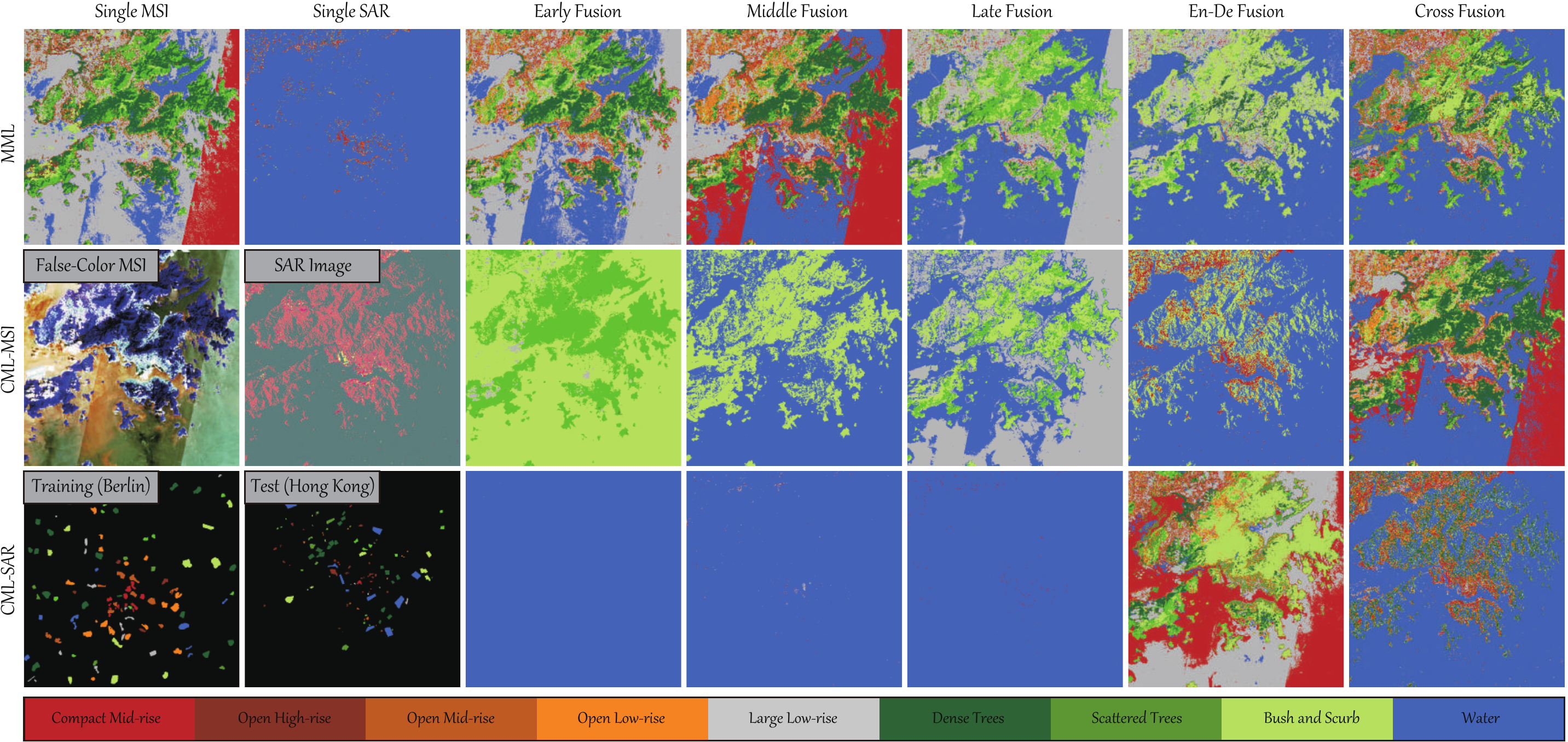}
		}
        \caption{Visualization of false-color MS and SAR images, the distribution of training and testing samples, and classification maps of different compared methods using FC-Nets on the MS-SAR LCZ data.}
\label{fig:CM_LCZ_FC}
\end{figure*}

\begin{figure*}[!t]
	  \centering
		\subfigure{
			\includegraphics[width=0.95\textwidth]{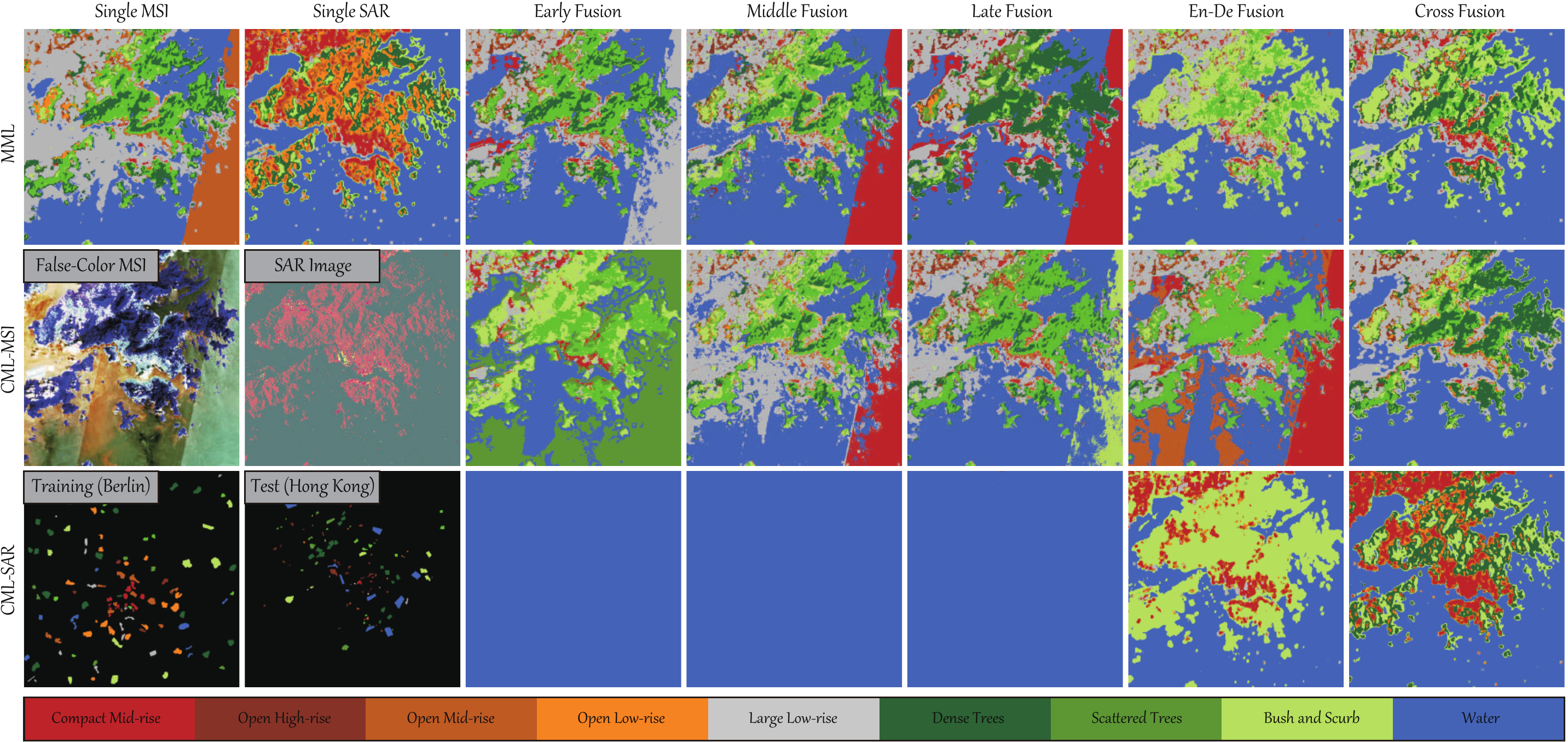}
		}
        \caption{Visualization of false-color MS and SAR images, the distribution of training and testing samples, and classification maps of different compared methods using CNNs on the MS-SAR LCZ data.}
\label{fig:CM_LCZ_CNN}
\end{figure*}

\subsubsection{Visual comparison}
Figs. \ref{fig:CM_HH_FC} and \ref{fig:CM_HH_CNN} visualize the classification maps of different networks for FC-Nets and CNNs, respectively, from which we have the following observations:
\begin{itemize}
    \item The MDL can provide a better solution than single modalities to reduce the errors in semantic labeling. Moreover, the compactness-based fusion approaches tend to generate more realistic classification maps.
    \item CNNs are able to achieve smoother inference results compared to FC-Nets by removing noisy pixels in classification maps.
    \item Multimodal data fusion is conducive to provide robust solutions against spectral variabilities, i.e., cloud cover in optical imaging, and alleviate the performance degradation by the means of other data sources (e.g., LiDAR).
    \item \textit{Cross fusion} module is capable of seeking out important visual, spectral, and other cues from highly complex materials lying in the image scene, thereby leading to a accurate reasoning result closer to the ground truth.
    \item In particular, the building-related types, e.g., \textit{Residential}, \textit{Commercial}, can be recognized better by \textit{en-de fusion} or \textit{cross fusion}, while some categories in the region covered by the cloud, such as \textit{Road}, \textit{Highway}, or \textit{Grass}, can be identified more accurately in CML-HSI using the compactness-based fusion strategy, due to more effective information transfer from LiDAR data.
\end{itemize}

\subsection{Result and Analysis on LCZ Data}
We evaluate the proposed MDL framework in a more challenging LCZ datasets (MS-SAR), where the main difficulties lie in two parts. On one hand, unlike the conventional LULC, LCZ defines more complex categories within a pixel at a very low spatial resolution, i.e., 100m. In this case, more diverse features and more powerful models are needed. On the other hand, due to completely different imaging mechanism, MS and SAR data are highly heterogeneous, posing a great challenge to the fusion of the two data in networks. Moreover, we select the datasets to investigate the network performance, e.g., transferability across cities (from Berlin to Hong Kong). As a result, the aforementioned factors can well explain that the classification performance in LCZ data (MSI and SAR) is inferior to that in Houston data (HSI and LiDAR), particularly in the result comparison between LiDAR data and SAR data when considering the CML's case.

\subsubsection{Quantitative comparison}
As listed in Tables \ref{tab:LCZ_MML_FC} and \ref{tab:LCZ_MML_CNN}, there is a basically consistent trend in performance gain with that on the HS-LiDAR Houston2013 datasets. In general, the results of using the proposed MDL-RS framework are much better than those of only using single modalities (averagely $10\%$ increase in \textit{OA}, \textit{AA}, and $\kappa$), while the CNN-based methods, as expected, exceed the FC-based ones at an increase of around 10\% in terms of all three indices. Despite so, we have to admit that our MDL-RS, to some extent, fails to recognize some materials, such as \textit{Open High-rise}, \textit{Open Low-rise}, \textit{Scattered Trees}, and \textit{Bush and Scrub}, especially in the CML's case. This may be due to imbalanced sample distribution and limited feature discrimination for the challenging LCZ categories. Moreover, the compactness-based networks outperform others remarkably at an improvement of over 5\% \textit{OA}, despite relatively low accuracies for certain categories obtained. It should be emphasized, however, that in CML those concatenation-based methods, i.e., \textit{early fusion}, \textit{middle fusion}, and \textit{late fusion}, are incapable of identifying ro even finding out certain materials in CML-MSI for example. It is much worse in CML-SAR, where all materials are recognized as \textit{Water}. On the contrary, either \textit{en-de fusion} or \textit{cross fusion} obtain better classification results, in particular, the latter brings a further improvement of almost 5\% \textit{OA} over the former.

\subsubsection{Visual comparison}
Similarly, visual differences between the classification maps of different networks are shown in Figs. \ref{fig:CM_LCZ_FC} and \ref{fig:CM_LCZ_CNN} for FC-Nets' and CNNs', respectively. In MML, the \textit{cross fusion} in our MDL-RS obtain a smoother and more detailed appearance in comparison with other fusion approaches, due to its use of cross learning strategy to eliminate the gap between modalities more effectively. A similar conclusion can be made in the \textit{en-de fusion} method with a slightly low accuracy compared to the \textit{cross fusion}. Moreover, the compactness-based methods are more robust against various image degradation, e.g., noise, stripe, etc., than others, as shown in Figs. \ref{fig:CM_LCZ_FC} and \ref{fig:CM_LCZ_CNN} where a direct evidence is given. For the CML's case, all pixels in the scene are assigned with the label of \textit{Water} using concatenation-based methods, which indicates a weak network's transferability across different modalities. It should be noted, however, that although the performance of the compactness-based methods is somewhat degraded in the CML's issue compared to that in the MML's, the transferability still remains desirable (see both Figs. \ref{fig:CM_LCZ_FC} and \ref{fig:CM_LCZ_CNN}).

\section{Conclusion}
In this paper, we propose a general MDL framework that consists of two subnetworks: \textit{Ex-Net} and \textit{Fu-Net}, aiming to provide a baseline solution for pixel-level RS image classification tasks using multimodal data. For this purpose, we investigate several different fusion strategies in networks with a focus on three questions: ``what'', ``where'', and ``how'' to fuse, as well as two kinds of feature extractors: FC-Nets and CNNs, which can be applicable to pixel-based and spatial-spectral classification, respectively. Apart from the well-studied MML problem, our MDL-RS framework also attempts to drive the research on the issue of CML that widely exists in practice but is less investigated. It should be emphasized, however, that we generalize four well-known fusion modules, e.g., \textit{early fusion}, \textit{middle fusion}, \textit{late fusion}, and \textit{en-de fusion} into the proposed MDL-RS framework, and additionally propose a novel fusion strategy: \textit{cross fusion} that not only performs better in MML but also is well applicable to CML. Experimental results conducted on two different multimodal RS datasets demonstrate the effectiveness and superiority of our MDL-RS networks compared to single modalities, and further the compactness-based fusion strategy is superior to the concatenation-based one as well, especially in the CML's case.
In summary,
\begin{itemize}
    \item In the ``what'' question, we mainly consider what kinds of modalities are used or fused in our MDL-RS framework. In this paper, we make the quantitative and visual comparison by using two different heterogeneous data, e.g., HS and LiDAR, MS and SAR, for RS image classification, and give a systematic and comprehensive analysis and discussion in the experimental section.
    \item In the ``where'' question, we investigate several well-known fusion modules, e.g., \textit{early fusion}, \textit{middle fusion}, \textit{late fusion}, which are corresponding to three different fusion locations in networks, respectively. By quantitative and qualitative assessment, we found that the \textit{middle fusion} and \textit{late fusion} tend to yield better classification results, particularly \textit{middle fusion}. It should be noted that as shown in Fig. \ref{fig:fusion}, the \textit{en-de fusion} and \textit{cross fusion} follow the same architecture as \textit{middle fusion}, that is, their fusion positions are also located in the ``middle'' of the network.
    \item In the ``how'' question, we also discuss two different fusion strategies, i.e., concatenation-based and compactness-based. The former is widely used in MML but usually fails to perform well in CML, while the latter, including \textit{en-de fusion} and newly-proposed \textit{cross fusion} show their superiority in blending multimodal features for either MML or CML setting.
\end{itemize}

However, the RS image classification extremely replies on the quality and amount of samples. Such dependence is stronger for DL-based models. To break the performance bottleneck in MDL, we will introduce weakly-supervised or self-supervised techniques into networks with better-designed fusion modules in the future work.

\section*{Acknowledgments}

The authors would like to thank the Hyperspectral Image Analysis group at the University of Houston and the IEEE GRSS DFC2013 for providing the CASI University of Houston datasets for the LULC classification task. The authors also would like to thank the IEEE GRSS DFC2017 for providing Sentinel-2 MS datasets for the LCZ classification task.

\bibliographystyle{ieeetr}
\bibliography{HDF_ref}

\begin{IEEEbiography}[{\includegraphics[width=1in,height=1.25in,clip,keepaspectratio]{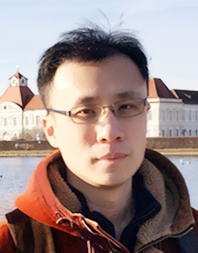}}]{Danfeng Hong}
(S'16--M'19) received the M.Sc. degree (summa cum laude) in computer vision, College of Information Engineering, Qingdao University, Qingdao, China, in 2015, the Dr. -Ing degree (summa cum laude) in Signal Processing in Earth Observation (SiPEO), Technical University of Munich (TUM), Munich, Germany, in 2019.

Since 2015, he worked as a Research Associate at the Remote Sensing Technology Institute (IMF), German Aerospace Center (DLR), Oberpfaffenhofen, Germany. Currently, he is a research scientist and leads a Spectral Vision working group at IMF, DLR, and also an adjunct scientist in GIPSA-lab, Grenoble INP, CNRS, Univ. Grenoble Alpes, Grenoble, France.

His research interests include signal / image processing and analysis, hyperspectral remote sensing, machine / deep learning, artificial intelligence and their applications in Earth Vision.
\end{IEEEbiography}

\vskip -2\baselineskip plus -1fil

\begin{IEEEbiography}[{\includegraphics[width=1in,height=1.25in,clip,keepaspectratio]{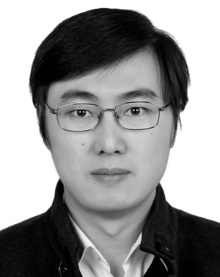}}]{Lianru Gao} (M'12--SM'18) received the B.S. degree in civil engineering from Tsinghua University, Beijing, China, in 2002, the Ph.D. degree in cartography and geographic information system from Institute of Remote Sensing Applications, Chinese Academy of Sciences (CAS), Beijing, China, in 2007.

He is currently a Professor with the Key Laboratory of Digital Earth Science, Aerospace Information Research Institute, CAS. He also has been a visiting scholar at the University of Extremadura, Cáceres, Spain, in 2014, and at the Mississippi State University (MSU), Starkville, USA, in 2016. His research focuses on hyperspectral image processing and information extraction. In last ten years, he was the PI of 10 scientific research projects at national and ministerial levels, including projects by the National Natural Science Foundation of China (2010-2012, 2016-2019, 2018-2020), and by the Key Research Program of the CAS (2013-2015). He has published more than 160 peer-reviewed papers, and there are more than 80 journal papers included by SCI. He was coauthor of an academic book entitled ``Hyperspectral Image Classification And Target Detection''. He obtained 28 National Invention Patents in China. He was awarded the Outstanding Science and Technology Achievement Prize of the CAS in 2016, and was supported by the China National Science Fund for Excellent Young Scholars in 2017, and won the Second Prize of The State Scientific and Technological Progress Award in 2018. He received the recognition of the Best Reviewers of the IEEE Journal of Selected Topics in Applied Earth Observations and Remote Sensing in 2015, and the Best Reviewers of the IEEE Transactions on Geoscience and Remote Sensing in 2017.
\end{IEEEbiography}

\vskip -2\baselineskip plus -1fil

\begin{IEEEbiography}[{\includegraphics[width=1in,height=1.25in,clip,keepaspectratio]{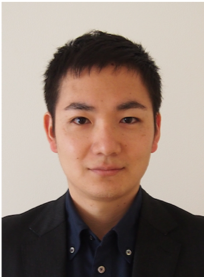}}]{Naoto Yokoya} (S'10--M'13) received the M.Eng. and Ph.D. degrees from the Department of Aeronautics and Astronautics, the University of Tokyo, Tokyo, Japan, in 2010 and 2013, respectively.

He is currently a Lecturer at the University of Tokyo and a Unit Leader at the RIKEN Center for Advanced Intelligence Project, Tokyo, Japan, where he leads the Geoinformatics Unit. He was an Assistant Professor at the University of Tokyo from 2013 to 2017. In 2015-2017, he was an Alexander von Humboldt Fellow, working at the German Aerospace Center (DLR), Oberpfaffenhofen, and Technical University of Munich (TUM), Munich, Germany. His research is focused on the development of image processing, data fusion, and machine learning algorithms for understanding remote sensing images, with applications to disaster management.

Dr. Yokoya won the first place in the 2017 IEEE Geoscience and Remote Sensing Society (GRSS) Data Fusion Contest organized by the Image Analysis and Data Fusion Technical Committee (IADF TC). He is the Chair (2019-2021) and was a Co-Chair (2017-2019) of IEEE GRSS IADF TC and also the secretary of the IEEE GRSS All Japan Joint Chapter since 2018. He is an Associate Editor for the IEEE Journal of Selected Topics in Applied Earth Observations and Remote Sensing (JSTARS) since 2018. He is/was a Guest Editor for the IEEE JSTARS in 2015-2016, for Remote Sensing in 2016-2020, and for the IEEE Geoscience and Remote Sensing Letters (GRSL) in 2018-2019.
\end{IEEEbiography}

\vskip -2\baselineskip plus -1fil

\begin{IEEEbiography}[{\includegraphics[width=1in,height=1.25in,clip,keepaspectratio]{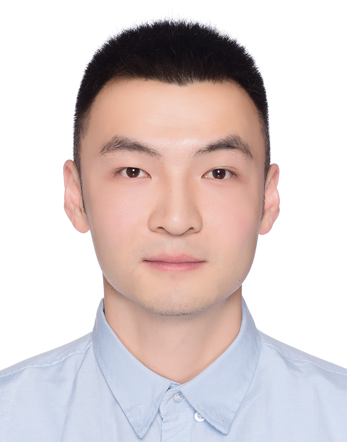}}]{Jing Yao} received the B.Sc. degree from Northwest University, Xi’an, China, in 2014. He is currently pursuing the Ph.D. degree with the School of Mathematics and Statistics, Xi’an Jiaotong University, Xi’an, China.

From 2019 to 2020, he is a visiting student at Signal Processing in Earth Observation (SiPEO), Technical University of Munich (TUM), Munich, Germany, and at the Remote Sensing Technology Institute (IMF), German Aerospace Center (DLR), Oberpfaffenhofen, Germany.

His research interests include low-rank modeling, hyperspectral image analysis and deep learning-based image processing methods.
\end{IEEEbiography}

\vskip -2\baselineskip plus -1fil

\begin{IEEEbiography}[{\includegraphics[width=1in,height=1.25in,clip,keepaspectratio]{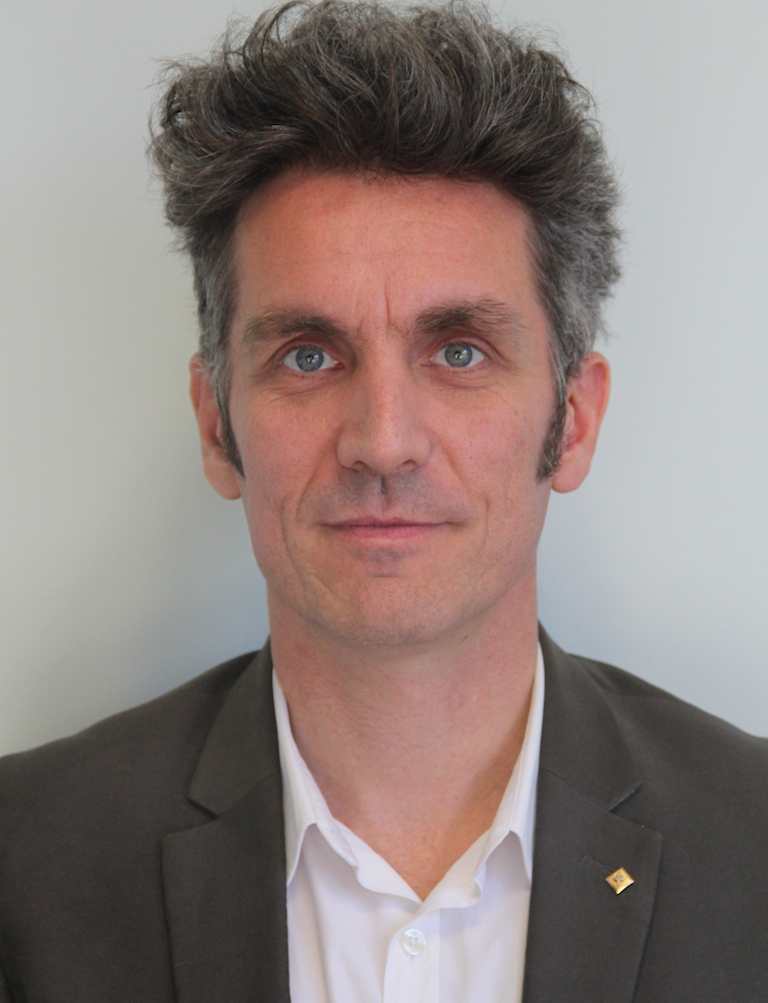}}]{Jocelyn Chanussot}
(M'04--SM'04--F'12) received the M.Sc. degree in electrical engineering from the Grenoble Institute of Technology (Grenoble INP), Grenoble, France, in 1995, and the Ph.D. degree from the Université de Savoie, Annecy, France, in 1998. Since 1999, he has been with Grenoble INP, where he is currently a Professor of signal and image processing. His research interests include image analysis, hyperspectral remote sensing, data fusion, machine learning and artificial intelligence. He has been a visiting scholar at Stanford University (USA), KTH (Sweden) and NUS (Singapore). Since 2013, he is an Adjunct Professor of the University of Iceland. In 2015-2017, he was a visiting professor at the University of California, Los Angeles (UCLA). He holds the AXA chair in remote sensing and is an Adjunct professor at the Chinese Academy of Sciences, Aerospace Information research Institute, Beijing.

Dr. Chanussot is the founding President of IEEE Geoscience and Remote Sensing French chapter (2007-2010) which received the 2010 IEEE GRS-S Chapter Excellence Award. He has received multiple outstanding paper awards. He was the Vice-President of the IEEE Geoscience and Remote Sensing Society, in charge of meetings and symposia (2017-2019). He was the General Chair of the first IEEE GRSS Workshop on Hyperspectral Image and Signal Processing, Evolution in Remote sensing (WHISPERS). He was the Chair (2009-2011) and  Cochair of the GRS Data Fusion Technical Committee (2005-2008). He was a member of the Machine Learning for Signal Processing Technical Committee of the IEEE Signal Processing Society (2006-2008) and the Program Chair of the IEEE International Workshop on Machine Learning for Signal Processing (2009). He is an Associate Editor for the IEEE Transactions on Geoscience and Remote Sensing, the IEEE Transactions on Image Processing and the Proceedings of the IEEE. He was the Editor-in-Chief of the IEEE Journal of Selected Topics in Applied Earth Observations and Remote Sensing (2011-2015). In 2014 he served as a Guest Editor for the IEEE Signal Processing Magazine. He is a Fellow of the IEEE, a member of the Institut Universitaire de France (2012-2017) and a Highly Cited Researcher (Clarivate Analytics/Thomson Reuters, 2018-2019).

\end{IEEEbiography}

\vskip -2\baselineskip plus -1fil

\begin{IEEEbiography}[{\includegraphics[width=1in,height=1.25in,clip,keepaspectratio]{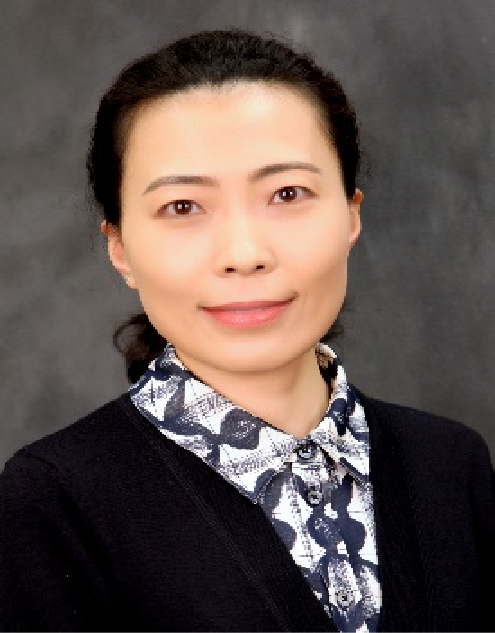}}]{Qian Du} (M'00--SM'05--F'18) received the Ph.D. degree in electrical engineering from the University of Maryland at Baltimore County, Baltimore, MD, USA, in 2000. She is currently the Bobby Shackouls Professor with the Department of Electrical and Computer Engineering, Mississippi State University, MS, USA. Her research interests include hyperspectral remote sensing image analysis and applications, pattern classification, data compression, and neural networks.

Dr. Du is a fellow of the SPIE-International Society for Optics and Photonics. She received the 2010 Best Reviewer Award from the IEEE Geoscience and Remote Sensing Society. She was the Co-Chair of the Data Fusion Technical Committee of the IEEE Geoscience and Remote Sensing Society from 2009 to 2013, and the Chair of the Remote Sensing and Mapping Technical Committee of the International Association for Pattern Recognition from 2010 to 2014. She has served as an Associate Editor of the IEEE JOURNAL OF SELECTED TOPICS IN APPLIED EARTH OBSERVATIONS AND REMOTE SENSING, the Journal of Applied Remote Sensing, and the IEEE SIGNAL PROCESSING LETTERS. She is the Editor-in-Chief of the IEEE JOURNAL OF SELECTED TOPICS IN APPLIED EARTH OBSERVATIONS AND REMOTE SENSING from 2016 to 2020.

\end{IEEEbiography}

\vskip -2\baselineskip plus -1fil

\begin{IEEEbiography}[{\includegraphics[width=1in,height=1.25in,clip,keepaspectratio]{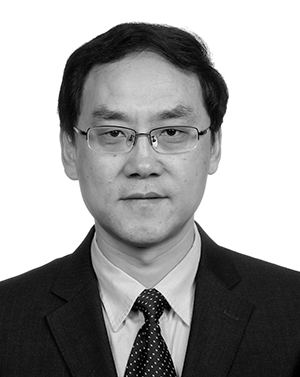}}]{Bing Zhang} (M'11--SM'12--F'19) received the B.S. degree in geography from Peking University, Beijing, China, in 1991, and the M.S. and Ph.D. degrees in remote sensing from the Institute of Remote Sensing Applications, Chinese Academy of Sciences (CAS), Beijing, China, in 1994 and 2003, respectively.

Currently, he is a Full Professor and the Deputy Director of the Aerospace Information Research Institute, CAS, where he has been leading lots of key scientific projects in the area of hyperspectral remote sensing for more than 25 years. His research interests include the development of Mathematical and Physical models and image processing software for the analysis of hyperspectral remote sensing data in many different areas. He has developed 5 software systems in the image processing and applications. His creative achievements were rewarded 10 important prizes from Chinese government, and special government allowances of the Chinese State Council. He was awarded the National Science Foundation for Distinguished Young Scholars of China in 2013, and was awarded the 2016 Outstanding Science and Technology Achievement Prize of the Chinese Academy of Sciences, the highest level of Awards for the CAS scholars.

Dr. Zhang has authored more than 300 publications, including more than 170 journal papers. He has edited 6 books/contributed book chapters on hyperspectral image processing and subsequent applications. He is the IEEE fellow and currently serving as the Associate Editor for IEEE Journal of Selected Topics in Applied Earth Observations and Remote Sensing. He has been serving as Technical Committee Member of IEEE Workshop on Hyperspectral Image and Signal Processing since 2011, and as the president of hyperspectral remote sensing committee of China National Committee of International Society for Digital Earth since 2012, and as the Standing Director of Chinese Society of Space Research (CSSR) since 2016. He is the Student Paper Competition Committee member in IGARSS from 2015-2019.
\end{IEEEbiography}

\end{document}